\documentclass[journal]{IEEEtran}
\ifCLASSINFOpdf
\else
\fi
\ifCLASSOPTIONcompsoc
 \usepackage[caption=false,font=normalsize,labelfont=sf,textfont=sf]{subfig}
\else
 \usepackage[caption=false,font=footnotesize]{subfig}
\fi
\usepackage{times}
\usepackage{epsfig}
\usepackage{graphicx}
\usepackage{amsmath}
\usepackage{amssymb}
\usepackage{multirow}
\usepackage{adjustbox}
\usepackage{tabu}
\usepackage[table]{xcolor}
\usepackage{soul}
\sethlcolor{lightgray}
\usepackage{bbold}

\usepackage{multicol}
\usepackage{booktabs}
\usepackage{hyperref}

\usepackage{algorithm}
\usepackage{algorithmicx}
\usepackage[noend]{algpseudocode}

\usepackage{xspace}
\makeatletter
\DeclareRobustCommand\onedot{\futurelet\@let@token\@onedot}
\def\@onedot{\ifx\@let@token.\else.\null\fi\xspace}
 
\def\ie{\emph{i.e}\onedot}

\def\etal{\emph{et al}\onedot}

\begin{document}
%
\title{Rethinking of Pedestrian Attribute Recognition:  \\ A Reliable Evaluation under Zero-Shot Pedestrian Identity Setting}
%
%

\author{Jian~Jia,~\IEEEmembership{Student~Member,~IEEE,}
        Houjing~Huang, ~\IEEEmembership{Student~Member,~IEEE,}
        Xiaotang~Chen,~\IEEEmembership{Member,~IEEE,}
        and~Kaiqi~Huang,~\IEEEmembership{Senior~Member,~IEEE}
\thanks{This work was supported in part by the National Natural Science Foundation of China (Grant No.61721004 and Grant No.61876181), the Projects of Chinese Academy of Science (Grant QYZDB-SSW-JSC006), the Strategic Priority Research Program of Chinese Academy of Sciences (Grant No. XDA27000000), and the Youth Innovation Promotion Association CAS. (Corresponding author: Kaiqi Huang.)}

\thanks{J. Jia and H. Huang are with the School of Artificial Intelligence, University of Chinese Academy of Sciences (UCAS), Beijing 100049, China, and also the Center for Research on Intelligent System and Engineering (CRISE),Institute of Automation, Chinese Academy of Sciences (CASIA), Beijing 100190, China (e-mail: jiajian2018@ia.ac.cn;houjing.huang@nlpr.ia.ac.cn)}

\thanks{X. Chen and K. Huang are with the Center for Research on Intelligent System and Engineering, Institute of Automation, Chinese Academy of Sciences, Beijing 100190, China, the University of Chinese Academy of Sciences, Beijing 100049, China. K. Huang is also with the CAS Center for Excellence in Brain Science and Intelligence Technology, Shanghai 200031, China (e-mail: xtchen@nlpr.ia.ac.cn; kqhuang@nlpr.ia.ac.cn)}
\thanks{Manuscript received April 19, 2005;}

}

%
%

\markboth{Journal of \LaTeX\ Class Files,~Vol.~14, No.~8, August~2015}%
{Shell \MakeLowercase{\textit{et al.}}: Bare Demo of IEEEtran.cls for IEEE Journals}
%



\maketitle

\begin{abstract}
Pedestrian attribute recognition aims to assign multiple attributes to one pedestrian image captured by a video surveillance camera. Although numerous methods are proposed and make tremendous progress, we argue that it is time to step back and analyze the status quo of the area. We review and rethink the recent progress from three perspectives. First, given that there is no explicit and complete definition of pedestrian attribute recognition, we formally define and distinguish pedestrian attribute recognition from other similar tasks. Second, based on the proposed definition, we expose the limitations of the existing datasets, which violate the academic norm and are inconsistent with the essential requirement of practical industry application. Thus, we propose two datasets, PETA\textsubscript{$ZS$} and RAP\textsubscript{$ZS$}, constructed following the zero-shot settings on pedestrian identity. In addition, we also introduce several realistic criteria for future pedestrian attribute dataset construction. Finally, we reimplement existing state-of-the-art methods and introduce a strong baseline method to give reliable evaluations and fair comparisons. Experiments are conducted on four existing datasets and two proposed datasets to measure progress on pedestrian attribute recognition.

\end{abstract}


%
\IEEEpeerreviewmaketitle

\section{Introduction}
%
%
%
%

\IEEEPARstart{P}{edestrian} attribute recognition (PAR) is to predict multiple attributes of 
pedestrian images as semantic descriptions in video surveillance, such as age, gender, and clothing. Recently, pedestrian attribute recognition has drawn increasing attention due to its great potential in real-world applications in video surveillance \cite{li2016richly, zhu2013pedestrian} and scene understanding \cite{li2019recurrent}, such as attribute classification, attribute retrieval \cite{li2018richly}, and person re-identification \cite{huang2020improve, yang2019towards}. Like many vision tasks, progress on pedestrian attribute recognition is significantly advanced by deep learning \cite{lecun2015deep}, and various methods \cite{li2015deepmar, yu2016weakly, liu2017hydraplus, wang2017attribute, li2018pose,li2018pedestrian, sarafianos2018deep, li2019visual, guo2019visual, lin2019improving} are proposed to improve the attribute recognition performance. All these methods focus on technical contributions to extract powerful attribute features. Nevertheless, two critical issues are still overlooked: What is the task of pedestrian attribute recognition? How to fairly evaluate the recognition performance of various methods in a realistic environment?

\begin{figure}[t]
\begin{center}
   \includegraphics[width=1\linewidth]{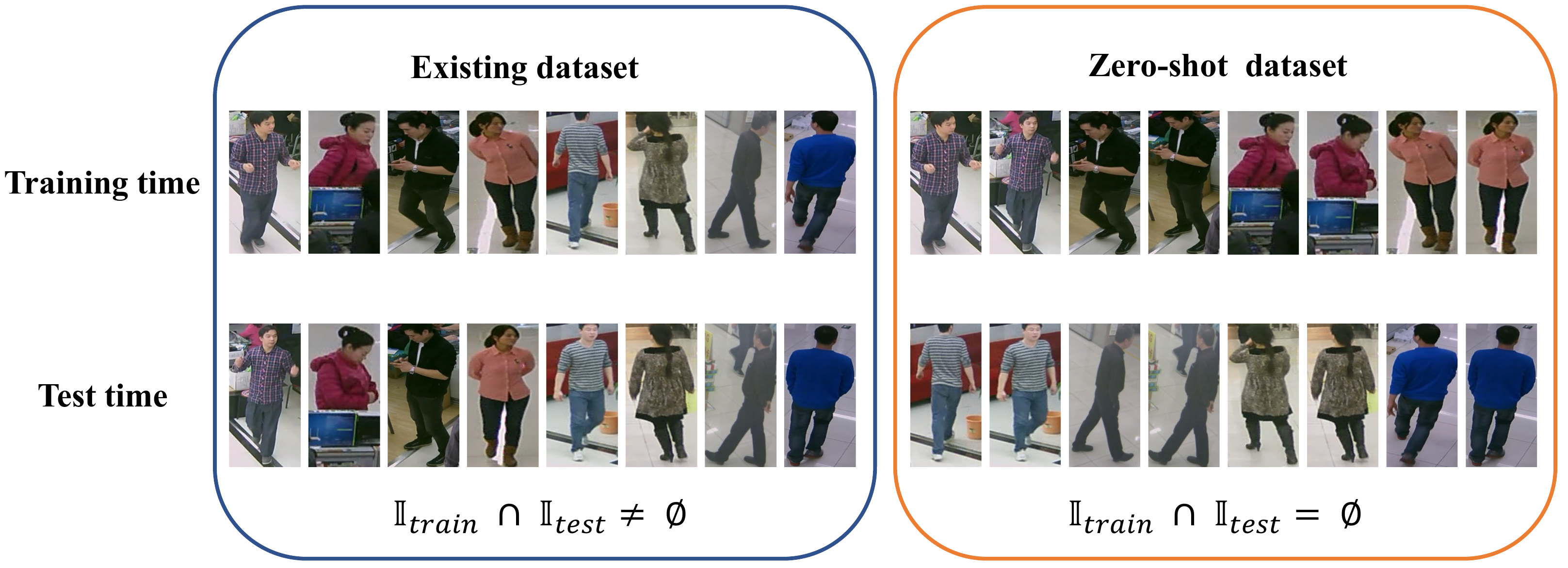}
\end{center}
   \caption{Comparison between existing dataset RAP and our proposed datasets RAP\textsubscript{ZS} of zero-shot pedestrian identity. In existing datasets RAP, plenty of pedestrian identities exists both in training time and test time, which contradicts with realistic application and academic criterion. In our proposed RAP\textsubscript{ZS}, we take the zero-shot setting of pedestrian identity between training time and test time, which can real evaluate the model performance and generalization.}
\label{fig:zeroshot_overview}
\vspace{-1em}
\end{figure}

Unlike the vision tasks such as detection \cite{li2018mixed}, segmentation\cite{gao2019ssap}, and tracking\cite{huang2019got}, although some previous works \cite{deng2014pedestrian, zhu2013pedestrian, li2018richly, liu2017hydraplus} focus on pedestrian attribute datasets, a clear and specific definition of pedestrian attribute recognition has yet been proposed. The ambiguous definition further makes datasets flawed and makes evaluation protocol unrealistic. Due to the ambiguous definition and questionable datasets, the performance of existing methods is overestimated, and it is difficult to quantify the actual progress in the pedestrian attribute recognition area. Therefore, we first introduce an explicit and specific definition of pedestrian attribute recognition to guide the dataset construction and performance evaluation. Second, to solve the limitations on existing datasets, several actionable criteria are proposed to construct realistic pedestrian attribute datasets PETA\textsubscript{ZS}, RAP\textsubscript{ZS} from the perspective of zero-shot pedestrian identity. We illustrate the comparison between existing pedestrian attribute datasets and zero-shot pedestrian attribute datasets in Figure \ref{fig:zeroshot_overview}. RAP and RAP\textsubscript{ZS} are taken as examples. Finally, we extensively evaluate recent state-of-the-art methods in depth on existing four popular datasets PETA \cite{deng2014pedestrian}, RAP1\cite{li2016richly}, RAP2\cite{li2018richly}, PA100K \cite{liu2017hydraplus} and proposed datasets PETA\textsubscript{ZS}, RAP\textsubscript{ZS}. 

\begin{figure}[t]
\begin{center}
   \includegraphics[width=1\linewidth]{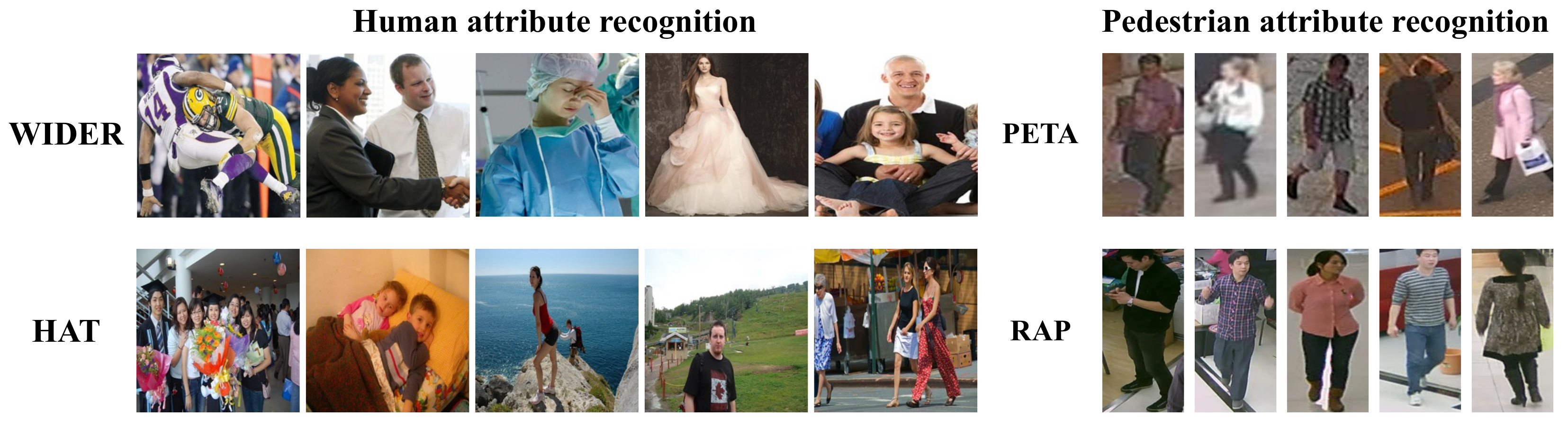}
\end{center}
   \caption{Comparison between human attribute recognition task and pedestrian attribute recognition task. For the human attribute recognition task, we take WIDER \cite{li2016human} and HAT \cite{sharma2011learning} datasets as examples. For the task of pedestrian attribute recognition, we take PETA \cite{deng2014pedestrian} and RAP \cite{li2018richly} datasets as examples. The ratio of images keeps 1:1 and 1:2 for two tasks, respectively.}
\label{fig:task_comprison}
\vspace{-1em}
\end{figure}

Great oaks from little acorns grow. As the ``acorn'' of pedestrian attribute recognition, an explicit and specific task definition is necessary for academic research and industry application. However, as far as we are concerned, there is no clear definition for pedestrian attribute recognition task since it is first proposed by Zhu \etal~ on APiS \cite{zhu2013pedestrian}, followed by PETA \cite{deng2014pedestrian}, RAP1 \cite{li2016richly}, PA100K \cite{liu2017hydraplus}, RAP2 \cite{li2018richly}, Market1501 \cite{lin2019improving}, and DukeMTMC \cite{lin2019improving}. Thus, we first introduce an explicit, specific, and complete definition to solid the base of pedestrian attribute recognition and distinguish it from other similar tasks like human attribute recognition. The core characteristic distinguishing pedestrian attribute recognition from other tasks lies in cropped pedestrian images, which are the bounding box results from the pedestrian detector or human annotator. Unlike images of the human attribute task, pedestrian images captured by surveillance cameras are low-resolution, and the entire pedestrian body is located in the center of the images. We illustrate the comparison between images of the human attribute recognition task and images of the pedestrian attribute recognition task in Fig. \ref{fig:task_comprison}.

Given a clear definition of pedestrian attribute recognition, we systematically evaluate existing datasets PETA \cite{deng2014pedestrian}, RAP1 \cite{li2016richly}, RAP2 \cite{li2018richly}, and PA100K \cite{liu2017hydraplus} in terms of academic norms and practical application. From the perspective of academic norms, due to the ambiguous task definition, previous works \cite{zhu2013pedestrian, li2016richly, li2018richly} randomly split the samples into the training set and test set without the validation set. All previous methods directly finetune hyperparameters in the test set, which violates the academic norms and makes method generalization ability improperly measured. Besides, as demonstrated in Fig. \ref{fig:zeroshot_overview}, the overlap of pedestrian identities between the training set and the test set makes some images of the test set almost identical to the images of the training set, except for little background and pose variation. This phenomenon is also called ``data leakage", which violates the academic norms. From the perspective of practical application, pedestrian attribute recognition is first proposed to provide auxiliary knowledge for applications, such as pedestrian tracking, person re-identification, and person retrieval \cite{zhu2013pedestrian}. In these applications, pedestrian (person) identities of the test set never appear in the training set, \ie,  zero-shot settings of pedestrian identities. However, plenty of pedestrian identities of the test set on PETA \cite{deng2014pedestrian} and RAP \cite{li2016richly} exist in the training set. Therefore, the standard of dividing the training set and the test set is inconsistent with the practical application. To solve the mentioned problem, we introduce a series of dataset splitting criteria and propose two new datasets PEAT\textsubscript{ZS} and RAP\textsubscript{ZS} based on original PEAT \cite{zhu2013pedestrian} and RAP \cite{li2018richly} following zero-shot settings of pedestrian identities.

After introducing the definition and proposed datasets, we reimplement and thoroughly evaluate current state-of-the-art methods, including MsVAA \cite{sarafianos2018deep}, VAC \cite{guo2019visual}, ALM \cite{tang2019Improving}, and JLAC \cite{tan2020relation}. The crux of the matter for methods of pedestrian attribute recognition is the unfair comparison and overestimated performance since experiments of these methods are conducted based on different experimental settings and tested on defective datasets. First, as listed in Table \ref{tab:baseline}, the baseline performance gap between the existing methods is significant, originating from various backbone networks and different experimental settings. Some methods are evaluated only on the part of three popular datasets. Second, performance of methods is overestimated because pedestrian identities of the test set appear in the training set. As demonstrated in Fig.\ref{fig:overlapped_perf_gap}, pedestrian identity overlap causes remarkably similar images in the training set and test set on PETA \cite{deng2014pedestrian}. The same phenomenon can also be noticed on RAP1 \cite{li2016richly} and RAP2 \cite{li2018richly}. Thus, to make a fair comparison between various methods and provide a reliable baseline performance, we reimplement several state-of-the-art methods and conduct exhaustive experiments on different settings.

The contributions of this paper are as follows:
\begin{itemize}
	\item For the definition, we answer two questions: What is pedestrian attribute recognition? What is the essential characteristic of pedestrian attribute recognition different from other similar tasks?
	\item For the datasets and evaluation protocol, we answer two questions: Can the existing datasets truly measure the performance of pedestrian attribute recognition methods? If not, on what kind of datasets can performance be measured appropriately?
	\item For the existing methods, we answer two questions: How does the existing method perform on existing datasets and proposed datasets in the same experimental settings? How much influence do different experimental factors have on performance?
\end{itemize}




\section{Related Work}
In this section, we give a comprehensive introduction to the progress of pedestrian attribute recognition from the perspective of datasets and methods. 

\subsection{Pedestrian Attribute Datasets}
The first pedestrian attribute dataset APiS (Attributed Pedestrians in Surveillance) was proposed by Zhu \etal~\cite{zhu2013pedestrian}, where 3,661 images with 11 binary and two multi-class attribute annotations are collected from three outdoor scenes and one indoor scene. To provide a well-aligned yet diverse dataset of pedestrians in realistic scenarios, Sudowe \etal~\cite{sudowe2015person} proposed PARSE-27k (Pedestrian Attribute Recognition on Sequences) dataset, where samples are captured by a moving camera in a city environment and annotated by ten attributes. Deng \etal~\cite{deng2014pedestrian} released a new large-scale dataset PETA (PEdesTrian Attribute), composed of 19,000 images with 61 binary and four multi-class attributes. These images are chosen and organized from ten publicly available small-scale datasets. To take viewpoints, occlusions, and body parts into consideration, Li \etal~\cite{li2016richly} built RAP (Richly Annotated Pedestrian) \footnote{We use RAP1 to indicate this version and use RAP2 to indicate the version published in \cite{li2018richly}} dataset with 41,585 pedestrian samples, captured from a real surveillance network with 26 video cameras at a shopping mall. Sixty-nine fine-grained attributes and three environmental factors (viewpoints, occlusion styles, and body parts) are annotated. PA-100K (Pedestrian Attribute) is the largest pedestrian attribute dataset introduced by Liu \etal~\cite{liu2017hydraplus} and composed of 100,000 images with 26 attributes. Based on RAP1 \cite{li2016richly}, Li \etal further enlarge the dataset size and propose RAP2 \cite{li2018richly} with 84,928 pedestrian samples and 72 attributes. In order to improve global feature representation of person re-identification,  Lin \etal \cite{lin2019improving} added attribute information as complementary local cues on Market1501 \cite{zheng2015scalable} and DukeMTMC \cite{ristani2016performance} datasets and constructed two new attribute datasets Market1501-Attribute and DukeMTMC-Attribute \cite{lin2019improving}.
PETA \cite{deng2014pedestrian}, RAP1 \cite{li2016richly}, PA100K \cite{liu2017hydraplus} are three popular pedestrian attribute recognition datasets used by mostly mainstream methods \cite{liu2017hydraplus, wang2017attribute, li2018richly, li2018pedestrian, li2018pose, li2019visual, li2019pedestrian, tang2019Improving,  tan2020relation}.

\subsection{Pedestrian Attribute Methods}
We review the existing methods from two aspects. One type of methods adopts auxiliary information like human parsing or human key points as prior or training supervision. The other type of methods only uses attribute labels as supervision.

For the methods utilizing human parsing and human key points as extra prior, the critical challenge is using the human knowledge to locate related attribute regions and improve attribute feature discrimination. Li \etal~\cite{li2018pose} utilized the CPM \cite{wei2016convolutional} model to generate human key points as coarse supervision to train the STN \cite{jaderberg2015spatial} models for regressing pedestrian pose. Inspired by human body structure information considered in SpindleNet \cite{zhao2017spindle}, Zhao \etal~\cite{zhao2018grouping} utilized human key points and adopted a body region proposal network to generate the head, upper body, and lower body regions. These body region proposals were pooled and input into the LSTM \cite{hochreiter1997long} as local features of the attribute groups. Li \etal~\cite{li2019pedestrian} combined the graph convolution network (GCN) with knowledge distillation from a pre-trained human parsing model to guide the attribute semantic relational learning. Tan \etal~\cite{tan2019attention} proposed the JLPLS-PAA framework composed of parsing attention, label attention, and spatial attention mechanism, where parsing attention mechanism utilizes pedestrian parsing network PSPNet to provide parsing results and aggregates features from different semantic human regions. 

For the methods without using extra human knowledge, mainstream methods focus on locating attribute-related regions in a weakly supervised manner, \ie, only attribute label are given. Yu \etal~\cite{yu2016weakly} introduced the WPAL network by concatenating multi-level features of flexible spatial pyramid pooling layers to utilize the information at different scales. Liu \etal~\cite{liu2017hydraplus} proposed HydraPlus-Net with multi-directional attention modules to exploit the multi-scale feature maps and enrich the final feature representations. Wang \etal~\cite{wang2017attribute} proposed the JRL network, which took attribute recognition task as a sequence prediction problem and adopted LSTM \cite{hochreiter1997long} to formulate the correlations between pedestrian attributes. Li \etal~\cite{li2019visual} proposed a graph-based reasoning framework including visual-to-semantic sub-network and semantic-to-visual sub-network. The two sub-networks are jointly optimized to capture spatial and semantic attribute relations. Tang \etal~\cite{tang2019Improving} proposed a new framework that combined FPN \cite{lin2017feature}, STN \cite{jaderberg2015spatial}, and SE \cite{hu2018squeeze} modules to enhance the attribute localization and make full use of pyramid feature maps. Tan \etal~\cite{tan2020relation} proposed the JLAC model, which includes attribute relation module and contextual relation module, to formulate attribute relation as a graph structure.

\section{Problem Definition}
In this section, we first introduce the complete definition of pedestrian attribute recognition in realistic settings. Then, a comparison between pedestrian attribute recognition and other task is introduced. Finally, we demonstrate the characteristics of pedestrian attribute recognition, which is the base of the dataset construction.

\subsection{Definition} \label{definition}
As the stone of the development of an area, an explicit and specific definition is indispensable. However, APiS \cite{zhu2013pedestrian}, PETA \cite{deng2014pedestrian}, RAP1 \cite{li2016richly}, RAP2 \cite{li2018richly}, and PA100k \cite{liu2017hydraplus} do not give a concrete definition of pedestrian attribute recognition yet. Some papers \cite{sarafianos2018deep, wu2020distraction} even confuse pedestrian attribute recognition with other tasks, \ie, human attribute recognition \cite{li2016human} and re-identification-related attribute recognition \cite{lin2019improving}. Thus, we first give a specific and complete definition for pedestrian attribute recognition as follows.

\noindent \textbf{Definition}   Given a training set of input-target pairs $\mathcal{D} = \{ (x_{i}, y_{i}), i=1,2,\dots,N \}$, pedestrian attribute recognition is a task assigning multiple attribute labels $y_{i} \in \{0, 1\}^{M}$ to one pedestrian image $x_{i}$ of the test set, where pedestrian identities $\mathbb{I}_{test}$ have never been seen in the training set, \ie, the zero-shot settings between identities of the training set and identities of the test set, $ \mathbb{I}_{train} \cap \mathbb{I}_{test} = \varnothing $.

According to the definition, the essence of pedestrian attribute recognition is to train a model on images of one part of pedestrians to generalize well on images of the other part of pedestrians. Specifically, the pedestrian identities $\mathbb{I}_{test}$ of the test set cannot appear in the pedestrian identities $\mathbb{I}_{train}$ of the training set. This principle is the foundation for the pedestrian attribute datasets to measure the generalization ability of the model correctly.

\subsection{Task Comparison}
There are two similar tasks compared to pedestrian attribute recognition: human attribute recognition and person re-identification-related ((ReID-related)) attribute recognition. 

Human attribute recognition is first proposed by Bourdev \etal~\cite{bourdev2011describing} on BAP (Berkeley Attributes of People) dataset and by Sharma\etal~\cite{sharma2011learning} on HAT (Human ATtributes) dataset. Followed by HAT, Li \etal~\cite{li2016human} proposes a large-scale human attribute recognition dataset WIDER, where images are collected from 30 events. Unlike pedestrian attribute recognition, where almost all images contain only one pedestrian and the pedestrian is located in the center of the image, images of human attribute recognition contain multiple people (Fig. \ref{fig:task_comprison}) with huge pose variations. From the image quality perspective, pedestrian attribute images collected from surveillance cameras are low-resolution, while images on human attribute datasets are high-resolution and selected from search engines. In addition, pedestrian attribute recognition can provide auxiliary information to help pedestrian tracking and re-identification, while human attribute recognition cannot. 

\begin{figure}[t]
\begin{center}
   \includegraphics[width=1\linewidth]{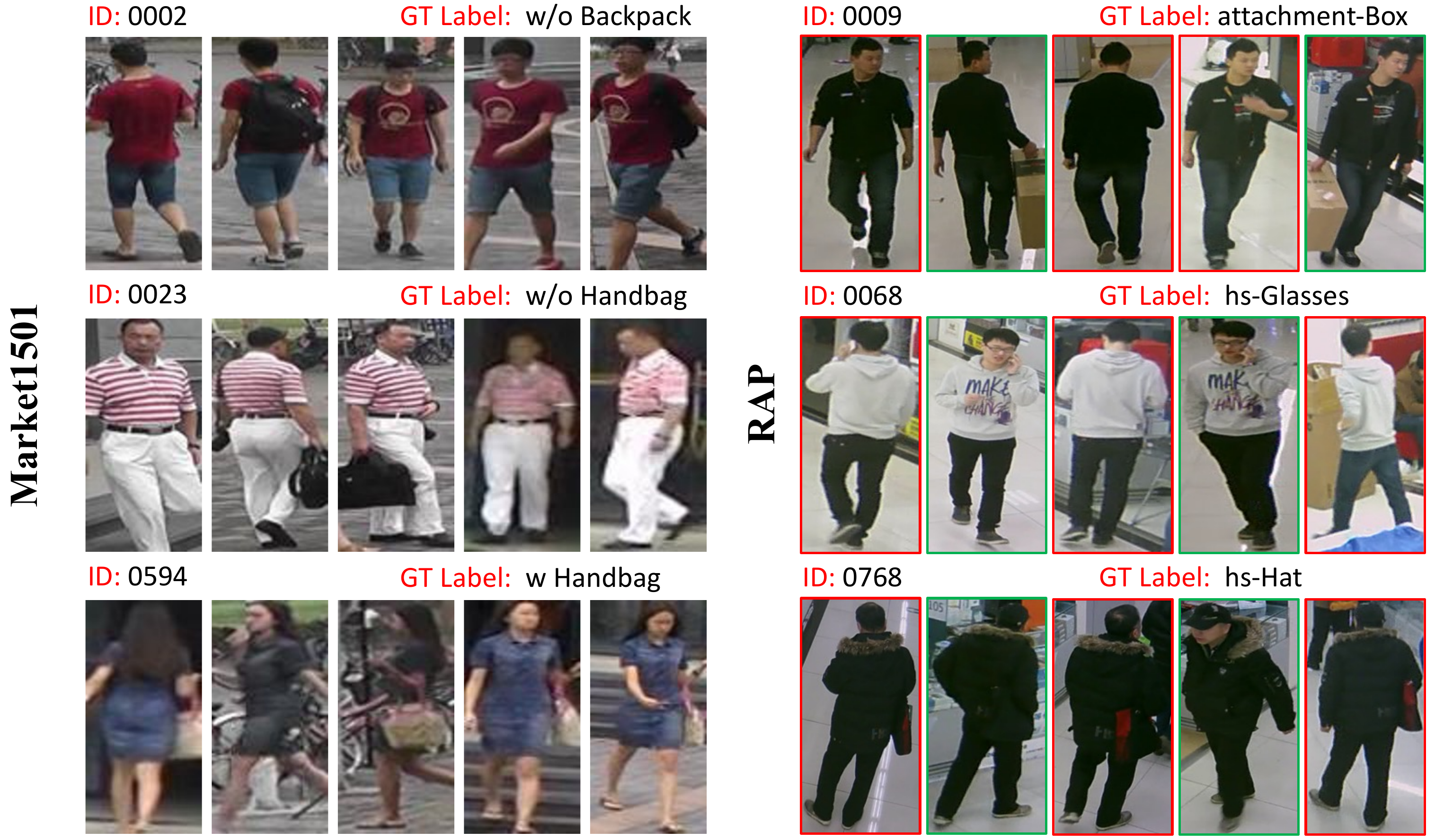}
\end{center}
   \caption{Comparison between ReID-related attribute recognition vs pedestrian attribute recognition. For the task of ReID-related attribute recognition and pedestrian attribute recognition, we take Market1501 \cite{lin2019improving} and RAP \cite{li2018richly} as examples, respectively. On Market1501 \cite{lin2019improving}, even if the attribute status changes, the same pedestrian identity is annotated with the same attribute labels, which is described as an identity-level annotation. On RAP \cite{li2018richly}, the same pedestrian identity may also have different attribute labels, depending on whether the attribute exists, which is described as an instance-level annotation. Images with \textbf{green} (\textbf{red}) boundaries are \textbf{positive} (\textbf{negative}) samples of the corresponding attribute. Same phenomenon can be also noticed on DukeMTMC \cite{lin2019improving}.}
\label{fig:reid_pedes}
\vspace{-1em}
\end{figure}

ReID-related attribute recognition refers to the attribute recognition task proposed by Lin \etal \cite{lin2019improving}, which is based on the ReID datasets and is specifically used to assist the ReID task. However, the existing ReID-related attribute datasets Market-1501 and DukeMTMC have two fatal flaws that make them unusable for pedestrian attribute recognition. First, annotations of pedestrian attribute datasets are at the instance level, while annotations of ReID-related attribute datasets are at the identity level. Datasets based on identity-level attribute annotation are unable to measure the model performance correctly. As shown in Fig. \ref{fig:reid_pedes}, we take three people on Market1501 as examples. For person ID0002, five images are labeled the ``without backpack'' attribute, regardless of the ``backpack" attribute in the second, third, and fifth images. This annotation standard is described as the identity-level annotation so that the attribute labels between all images of the same person are consistent. Nevertheless, this annotation leads to plenty of wrong labels, making the model unable to learn representative attribute features. As a result, model performance tested on these annotations is unreliable. On the contrary, for pedestrian attribute datasets, different images of the same pedestrian are annotated with different labels based on whether the attribute appears, which is described as the instance-level attribute annotation. For example, for pedestrian ID0068, the first, third, and fifth images are labeled as negative samples of the ``hs-Glasses'' attribute, while the second and fourth images are labeled as positive samples. Second, 17 of the 27 attributes in Market1501 and 17 of the 23 attributes in DukeMTMC belong to color attributes, which is not adopted for model evaluation \cite{li2016richly, li2018richly, li2019recurrent, li2019recurrent, tang2019Improving, liu2017hydraplus, sarafianos2018deep}. Besides, color attributes can be directly inferred by pixel values of RGB channels without model learning.

In a word, because of the different image collection scenes and the different annotation standards, pedestrian attribute recognition is quite distinct from human attribute recognition and ReID-related attribute recognition. Therefore, the following sections mainly focus on pedestrian attribute recognition and three widely used datasets PETA, RAP, and PA100K.

\section{Datasets}
In this section, we aim at answering two questions: Can the existing datasets truly measure the performance of pedestrian attribute recognition methods? If not, on what kind of datasets can performance be measured appropriately? Therefore, we first analyze the existing datasets and uncover the limitations. Then, we give a series of criteria to guide the pedestrian attribute dataset construction according to the task definition. Finally, according to the criteria, we propose two new pedestrian attribute datasets PEAT\textsubscript{ZS} and RAP\textsubscript{ZS}, following zero-shot settings on pedestrian identity.

\subsection{Flaws in existing datasets}

As defined in Section \ref{definition}, the critical requirement for pedestrian attribute datasets is to evaluate the model generalization ability correctly. However, due to the two limitations, this requirement is ignored by many previous dataset works. One limitation is the manner of collecting pedestrian images.

Pedestrian images of existing large-scale pedestrian attribute datasets are captured by surveillance video cameras in a dense sampling manner, such as 15 frames per second in RAP1 \cite{li2016richly} and RAP2 \cite{li2018richly}, which results little variation between adjacent frames of the same pedestrian.  The other limitation is the manner of splitting the training set and test set. Most existing datasets PETA \cite{deng2014pedestrian}, RAP1 \cite{li2016richly}, and RAP2 \cite{li2018richly} assign pedestrian images randomly into the training set and the test set without considering the pedestrian identities. As a result, the limitations in image collection manner and dataset splitting manner lead to the highly similar images between the training set and the test set, except for negligible background and pose variation as illustrated in Fig. \ref{fig:zeroshot_overview}. This problem is often referred to as ``data leakage " and makes it impossible to measure model generalization ability on most existing datasets, which contradicts academic norms and practical application objectives. Although PA-100K assigns all samples of one pedestrian to one of the training set and test set to avoid this problem, detailed explanations and underlying reasons are not given.

\begin{figure}[t]
\centering
\includegraphics[width=0.8\linewidth ]{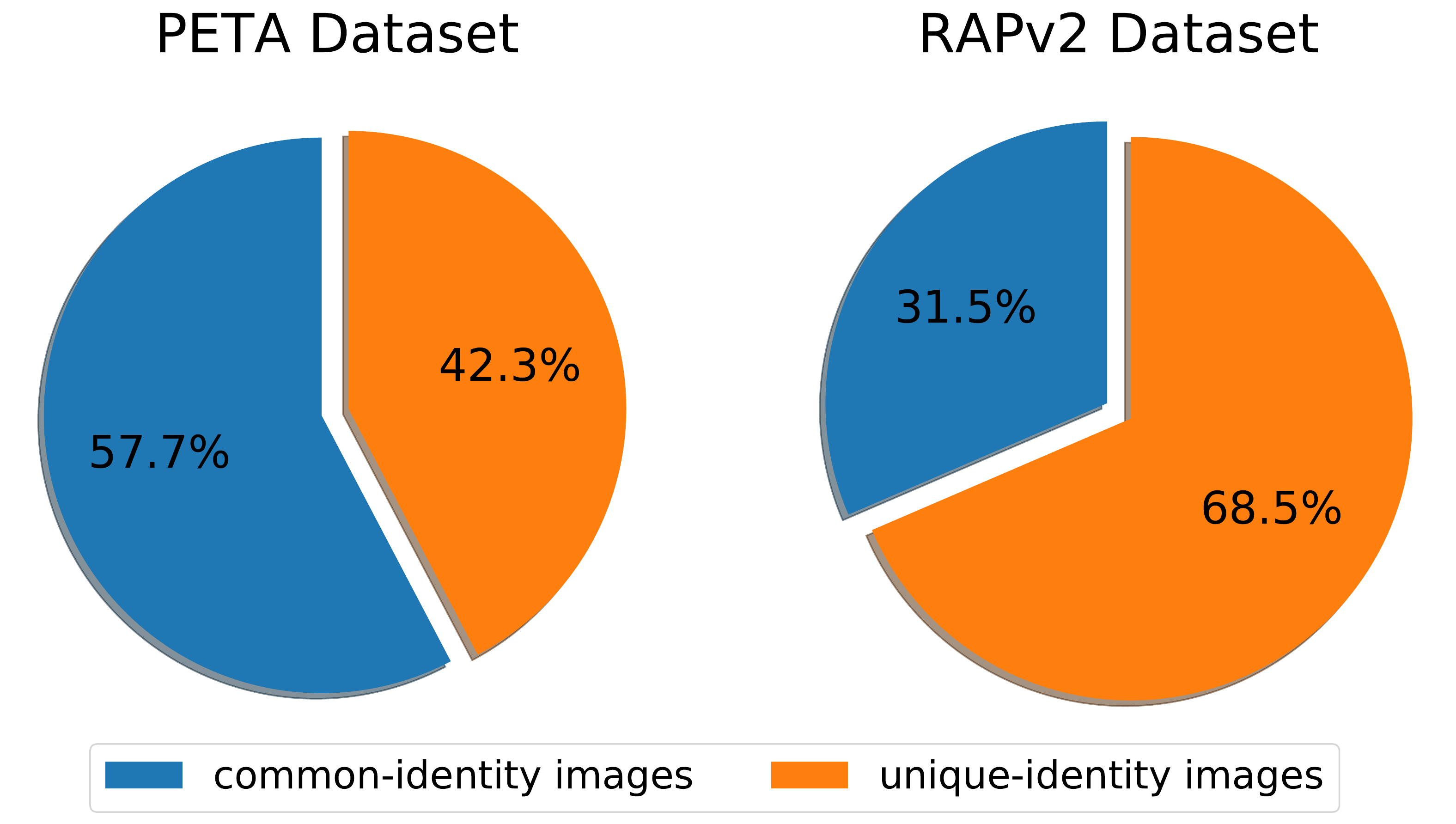}
\caption{The proportions of common-identities images in the test sets of PETA and RAP2. The proportion of common-identity images on RAP2 test set is at least 31.5\%, due to some test images are not labeled with pedestrian identity.}
\label{fig:common_image}
\vspace{-1em}
\end{figure}

To further confirm the severity of the problem, we statistic the proportion of common-identity images in the test set on PETA and RAP2. We use ``common-identity images" to represent the images of pedestrian identities that appear in both the training set and test set. In contrast, ``unique-identity images" are denoted as the images of pedestrian identities that appear only in the test set. As shown in Fig. \ref{fig:common_image}, 57.7\% and 31.5\% of test images have highly similar counterparts of the same identity in the training set of PETA and RAP2, respectively. Although there are also plenty of similar images between the training set and test set in RAP1, we can not get the accurate proportion of common-identity images in the test set, due to the lack of pedestrian identity labels.

\begin{figure*}[ht]
\centering
\subfloat[MsVAA]{
\includegraphics[width=0.5\linewidth]{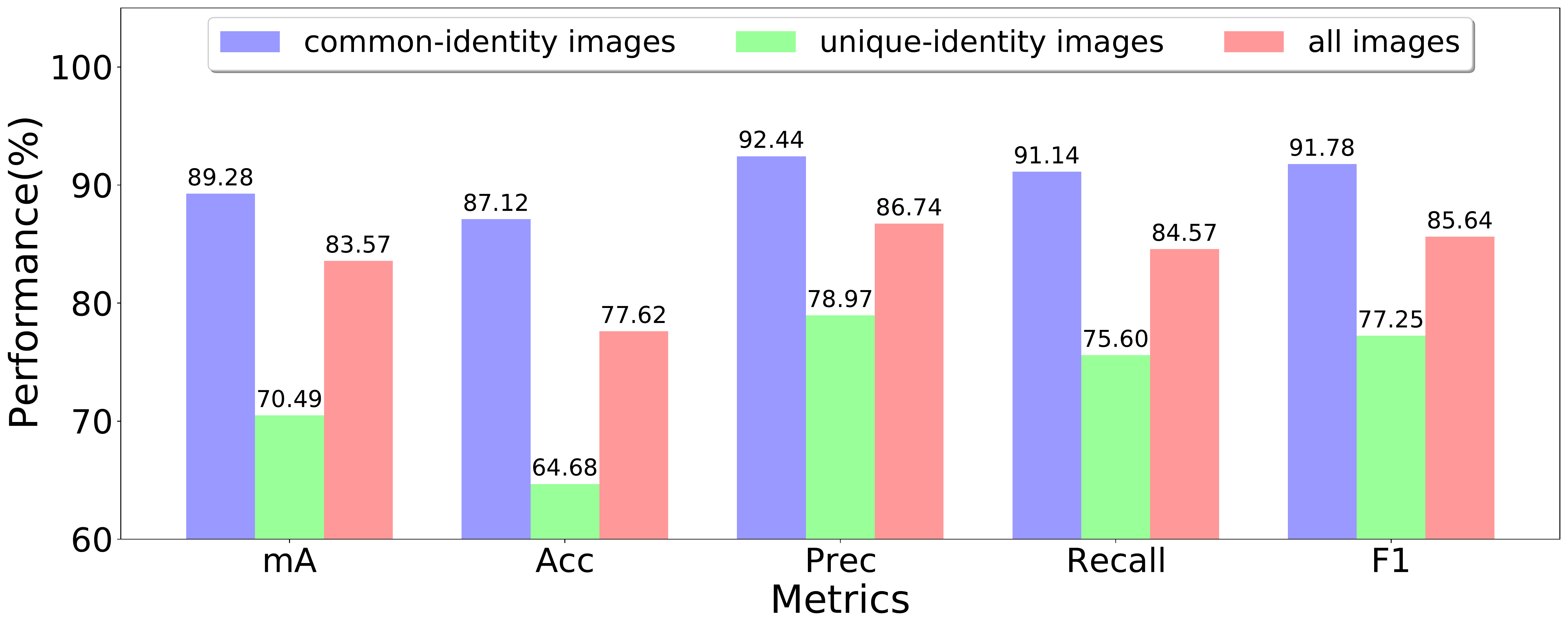}}
\subfloat[VAC]{
\includegraphics[width=0.5\linewidth]{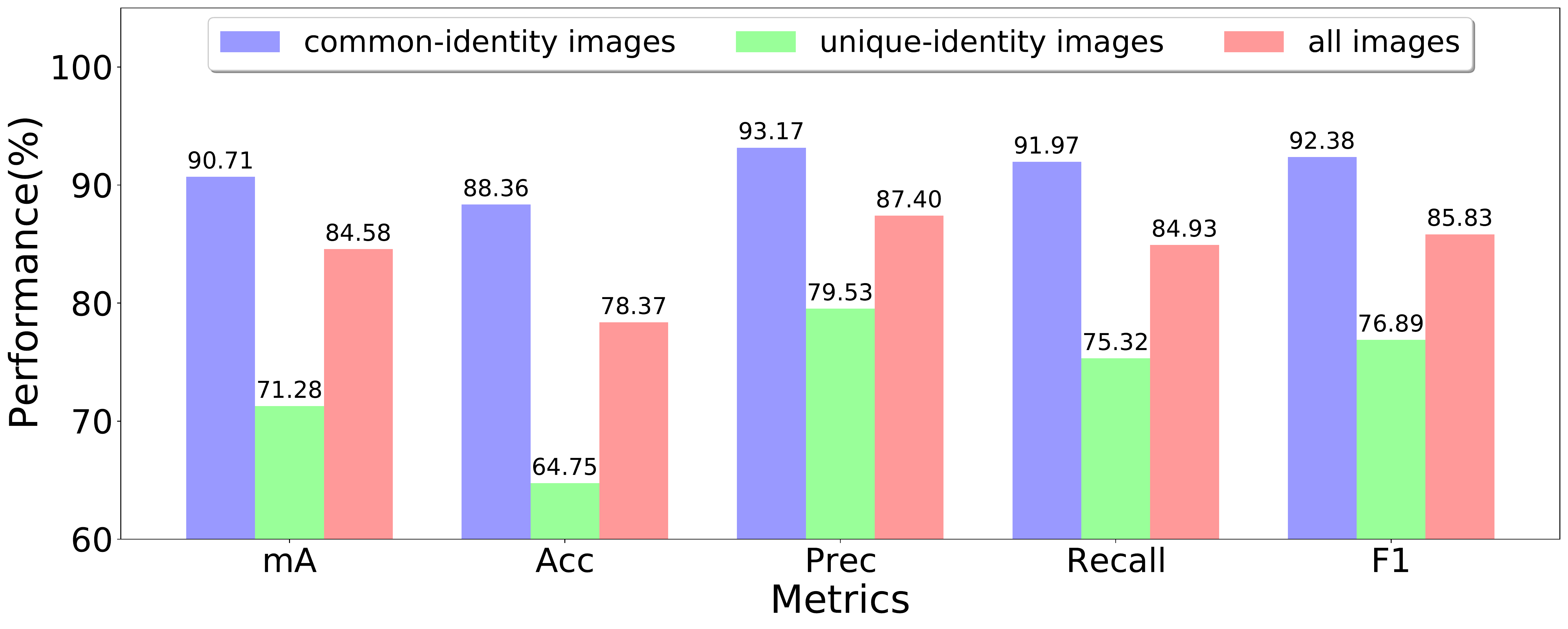}} \\
\subfloat[ALM]{
\includegraphics[width=0.5\linewidth]{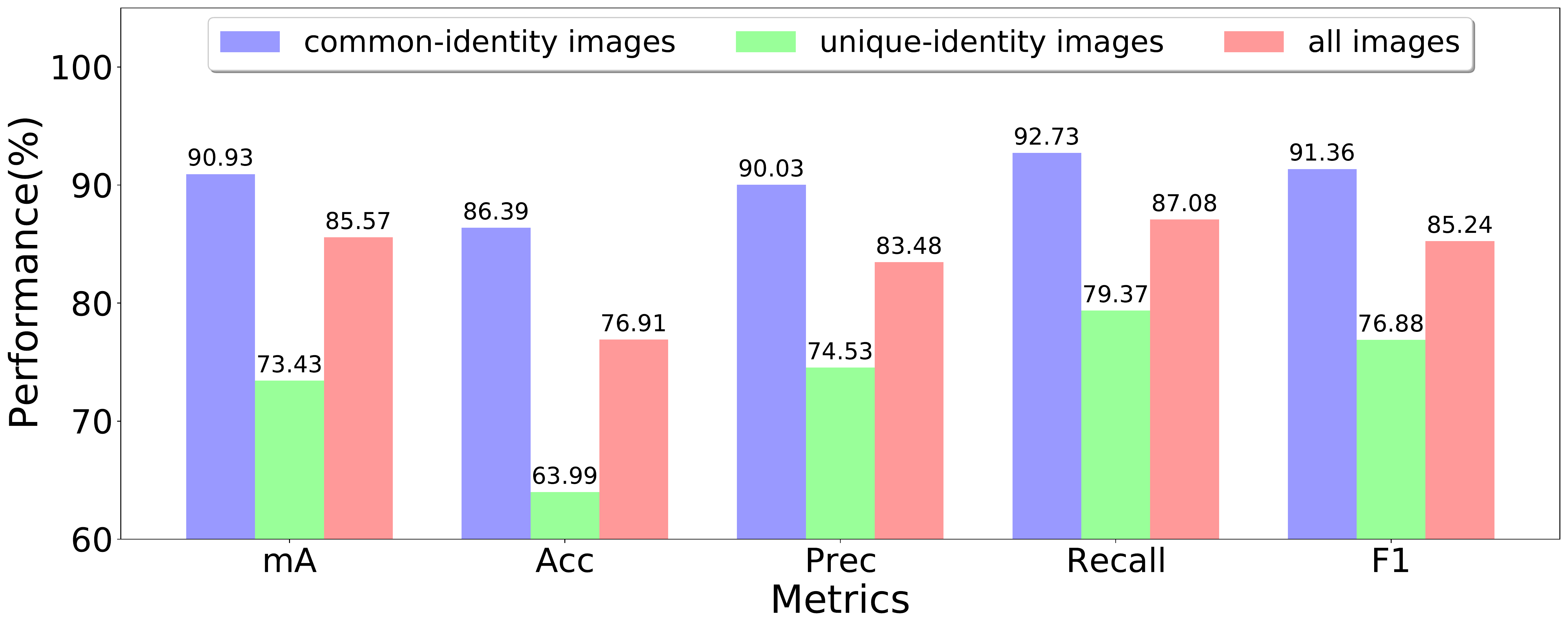}}
\subfloat[JLAC]{
\includegraphics[width=0.5\linewidth]{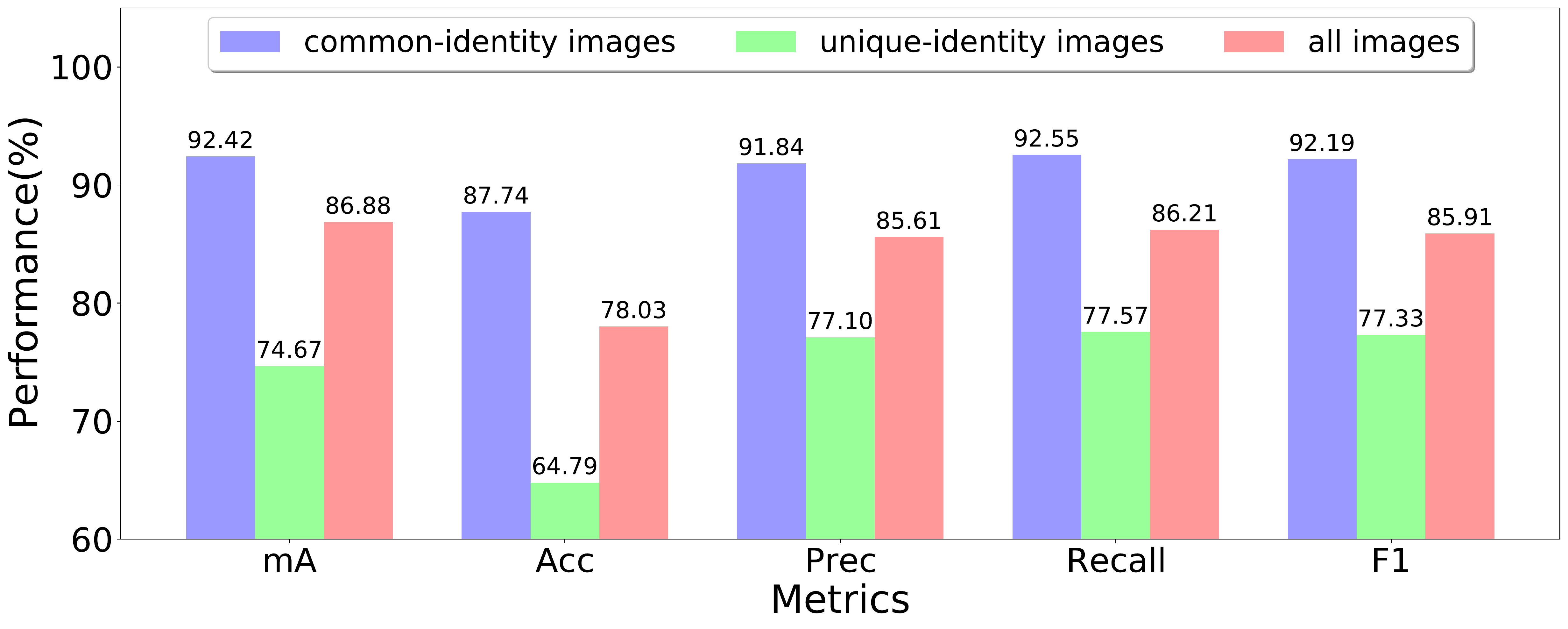}}
\caption{Performance of MsVAA \cite{sarafianos2018deep} and VAC \cite{guo2019visual} methods on common-identity images, unique-identity images and all images of PETA test set. We observe there is a big gap between common-identity images and unique-identity images as well as between unique-identity images and all images of test set. The remarkable performance difference shows the irrationality of existing datasets. Similar phenomenon can be observed on RAPv2 as well.}
\label{fig:overlapped_perf_gap}
\vspace{-1em}
\end{figure*}

Since plenty of test images have been seen in the training stage, recognition performance in the test set cannot reflect the method generalization ability. To give an intuitive demonstration, we reimplement four state-of-the-art methods MsVAA \cite{sarafianos2018deep}, VAC \cite{guo2019visual}, ALM \cite{tang2019Improving}, and JLAC \cite{tan2020relation}, and report their performance of common-identity images, unique-identity images, and all images on the test set. A significant performance gap between performance in common-identity images and performance in unique-identity images for four methods is shown in Fig.~\ref{fig:overlapped_perf_gap}. The phenomenon validates the argument that performance in the test set of existing datasets cannot reflect the generalization ability of methods. More importantly, the test set of the existing datasets, \ie, PETA, RAP1, and RAP2, overestimate the model performance and mislead the model evaluation.

\subsection{Splitting Criterion} \label{criterion}

To solve the mentioned flaws in existing datasets, we propose a series of criteria to guide the pedestrian attribute dataset construction as follows.

\noindent \textbf{Criteria on Pedestrian Attribute Dataset Construction:} 

\begin{enumerate}
\item $\mathcal{I}_{all} =\mathcal{I}_{train} \cup \mathcal{I}_{valid} \cup \mathcal{I}_{test}$, \\ $ |\mathcal{I}_{train}| : | \mathcal{I}_{valid}| : | \mathcal{I}_{test}|$ $\approx$  $ 3 : 1 : 1 $ . \label{crit1}
\item $\mathcal{I}_{train} \cap \mathcal{I}_{valid} = \varnothing$, $ \mathcal{I}_{train} \cap \mathcal{I}_{test} = \varnothing $, $\mathcal{I}_{valid} \cap \mathcal{I}_{test} = \varnothing$. \label{crit2}
\item $||\mathcal{I}_{valid}| - |\mathcal{I}_{test}|| < |\mathcal{I}_{all}| \times  T_{id}$ . \label{crit3}
\item $|N_{valid} - N_{test}| < T_{img}$ . \label{crit4}
\item $|R_{valid}^{i} - R_{test}^{i}| < T_{attr}, i = 1, 2, \dots, M$ . \label{crit5}
\end{enumerate}
where $\mathcal{I}_{train}$, $\mathcal{I}_{valid}$, $\mathcal{I}_{test}$ denotes the pedestrian identity set of the training set, validation set, and test set respectively. $|\cdot|$ denotes the set cardinality. $N_{valid}$, $N_{test}$ indicates the number of samples of validation set and test set. $R_{valid}^{i}$, $R_{test}^{i}$ is the positive ratio of $i$-th attribute in the validation set and test set. $T_{id}$, $T_{img}$, $T_{attr}$ is the threshold to control the difference of identity number, image number, and positive number between the validation set and test set. \textbf{Criteria~\ref{crit1}}, \textbf{Criteria~\ref{crit3}}, and \textbf{Criteria~\ref{crit4}} are used to provide an effective and reliable validation set, which is necessary for tuning method hyper-parameters. We argue that tuning hyper-parameters on test classes to improve the pedestrian attribute recognition performance violates the zero-shot assumption \cite{xian2018zero}. \textbf{Criteria~\ref{crit2}} is proposed to follow the zero-shot settings on pedestrian identity. In addition, \textbf{Criteria~\ref{crit5}} is proposed to control the attribute distribution gap between the validation set and the test set.

\subsection{Existing Datasets and Proposed Datasets}

\begin{algorithm}[t]
    \caption{Pseudocode of Splitting Search Algorithm in a Numpy-like style. }
  \label{alg:splitting_search}
\begin{algorithmic}[1]
  \footnotesize
  \Require Pedestrian dataset $\mathcal{D} = \{x_{i}, y_{i}, Id_{j} | i=1, \dots, N, j=1, \dots, K\}$, where $\mathcal{I} = \{Id_{j}|j=1, \dots, K\}$ denotes the pedestrian identity set, $\mathcal{X} = \{x_{i}|i=1, \dots, N \}$, $\mathcal{Y} = \{y_{i} | i=1, 2, \dots, N \}$, and $y_{i} \in \{0, 1\}^{M}$. Id2Idx = Dict(). The threshold of pedestrian identity number difference, attribute positive radio difference, and images number difference is set $T_{id} = 50$, $T_{attr} = 0.03$, $T_{img} = 300$ as default respectively.
  \For{$k = 1$ to $K$}
    \State $\text{Id2Idx}[k] = \{x_{i}, y_{i}, Id_{j} | Id_{j} = k\}$ 
  \EndFor
  \While{True}
    \State $K_{train} = \text{Random.Randint}([ \frac{3}{5}\cdot K - T_{id}, \frac{3}{5} \cdot K + T_{id}]) \label{get_train_id_start}$ 
    \State $\mathcal{I}_{train} = \text{Random.Choice}(\mathcal{I}, K_{train}, \text{replace=False})  $
    \State $K_{mid} = (K - K_{train}) / 2$
    \State $K_{valid} = \text{Random.Randint}([K_{mid} - T_{id}, K_{mid} + T_{id}]) \label{id_cons}$ 
    \State $\mathcal{I}_{valid} = \text{Ramdom.Choice}(\mathcal{I} - \mathcal{I}_{train}, K_{valid}, \text{replace=False})$
    \State $K_{test} = K - K_{train} - K{valid}$
    \State $\mathcal{I}_{test} = \mathcal{I} - \mathcal{I}_{train} - \mathcal{I}_{valid}$
    \State $\mathcal{D}_{train}=\mathcal{D}_{valid}=\mathcal{D}_{test}= \varnothing$
  \For{$k = 1$ to $K$}
    \If{$k \in \mathcal{I}_{train}$}
        \State $\mathcal{D}_{train} \leftarrow \mathcal{D}_{train} \cup \text{Id2Idx}[k]$
    \ElsIf{$k \in \mathcal{I}_{valid}$}
        \State $\mathcal{D}_{valid} \leftarrow \mathcal{D}_{valid} \cup \text{Id2Idx}[k]$
    \Else
        \State $\mathcal{D}_{test} \leftarrow \mathcal{D}_{test} \cup \text{Id2Idx}[k]$ \label{get_train_id_end}
    \EndIf
  \EndFor
  
  \If{$ ||\mathcal{D}_{test}| - |\mathcal{D}_{val}|| > T_{img} $} \label{img_cons}
    \State continue
  \EndIf
  
  \State Collect label set $\mathcal{Y}_{train}$, $\mathcal{Y}_{valid}$, $\mathcal{Y}_{test}$ from dataset $\mathcal{D}_{train}$, $\mathcal{D}_{valid}$, $\mathcal{D}_{test}$ respectively.
  \State Compute positive ratio $R_{train}$, $R_{valid}$, $R_{test} \in (0, 1)^{M}$ of each attribute in training set, validation set and test set.
  \If{($R_{train} - R_{valid} < T_{attr}$).all() and ($R_{train} - R_{test} < T_{attr}$).all()} \label{contr_same}
    \State \Return{$\mathcal{D}_{train}$, $\mathcal{D}_{test}$, $\mathcal{D}_{val}$} 
  \EndIf
  \EndWhile
\end{algorithmic}
\end{algorithm}

PETA \cite{deng2014pedestrian}, RAP1 \cite{li2016richly}, PA100K \cite{liu2017hydraplus} are three widely adopted pedestrian attribute datasets in recent state-of-the-arts methods. Although RAP2 \cite{li2018richly} is expanded on RAP1 and has more samples, all methods conduct experiments on RAP1 to make a fair comparison with previous methods. Except for PA100K, PETA, RAP1, and RAP2 adopt random splitting criterion to construct the training data and test set, which violates the zero-shot settings on pedestrian identity so that performance cannot be correctly evaluated in the test set. Considering that no pedestrian identity label is available in RAP1, we propose PETA \textsubscript{ZS} and RAP \textsubscript{ZS} datasets based on existing PETA and RAP2 datasets.

\begin{figure}[t]
\centering
\includegraphics[width=1\linewidth]{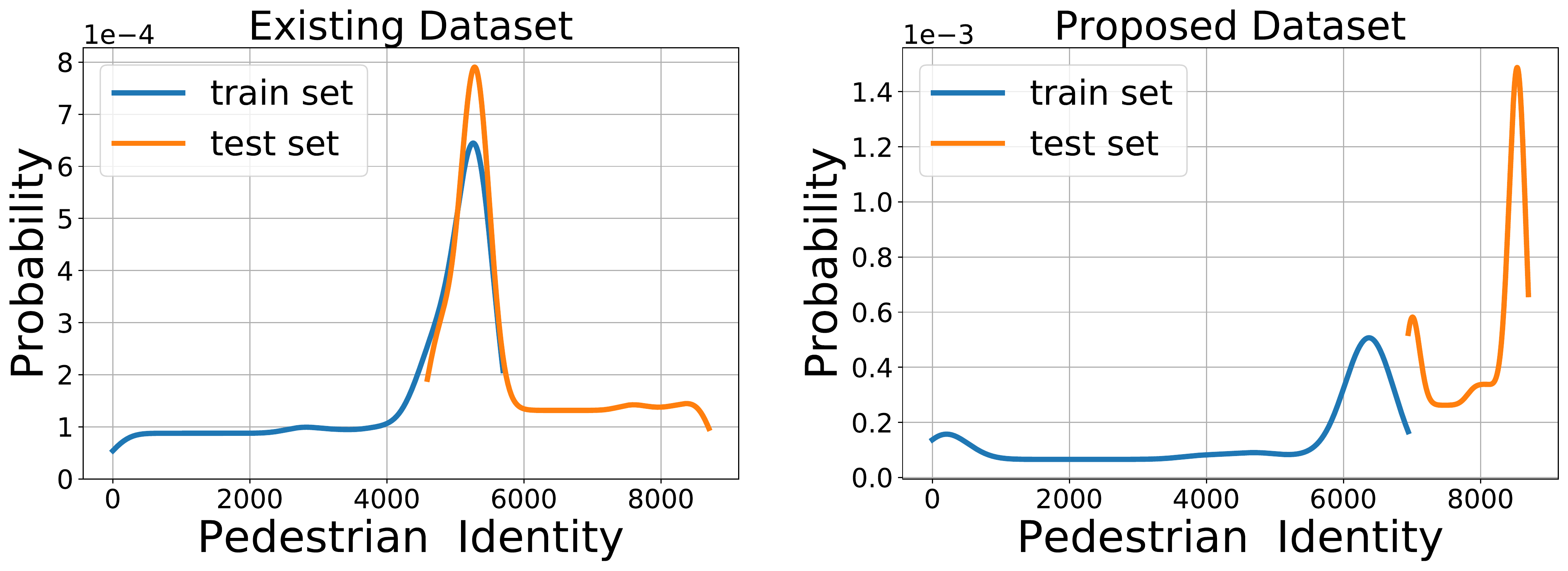}
\caption{Pedestrian identity distribution on PETA and PETA\textsubscript{$zs$}. X-axis indicates the pedestrian identity number and Y-axis is the proportion of corresponding images of the pedestrian identity. There are 1106 common-identities between the training set and the test set, accounting for 19.42 percent of the pedestrian identities in the training set (55.27 percent of the image) and 26.91 percent of the pedestrian identities in the test set (57.70 percent of the image). Our proposed dataset PETA\textsubscript{$ZS$} solves the problem by completely separating the pedestrian identities of the test set from the identities of the training set.}
\label{fig:overlap_dataset}
\vspace{-1em}
\end{figure}



In order to avoid the influence of artificially splitting the dataset, we propose the splitting search algorithm to find suitable dataset partitions under the guidance of the criteria on pedestrian attribute dataset construction. The full algorithm is provided in Algorithm \ref{alg:splitting_search}. Besides following the zero-shot settings of pedestrian identity (Algorithm \ref{alg:splitting_search}, line \ref{get_train_id_start}-\ref{get_train_id_end}), the attribute distribution of training set, validation set, and test set is approximately consistent and constrained by threshold $T_{attr}$ (Algorithm \ref{alg:splitting_search}, line \ref{contr_same}). To construct an effective validation set for performance measurement and hyper-parameters fine-tuning, we use the threshold $T_{img}$ and $T_{id}$ to separately limit the difference in the number of images and the number of identities between the validation set and the test set.  (Algorithm \ref{alg:splitting_search}, line \ref{id_cons} and line \ref{img_cons}).

Targeting at providing a reliable dataset partition and avoid the influence from random noise, we conduct the search algorithm five times on each dataset PETA, RAP2 and get five versions of PETA\textsubscript{ZS}, RAP\textsubscript{ZS}. The dataset information and performance of the baseline method (described in Section \ref{method}) on these datasets are reported in Table \ref{tab:five_version}. For PETA\textsubscript{ZS}, we keep all images and annotated attributes unchanged. For RAP\textsubscript{ZS}, we only keep images with pedestrian identity labels and remove the ``Age46-60" attribute because there is no positive sample of the ``Age46-60" attribute. From the detailed dataset information and experiment performance of five versions, we can conclude that there is no apparent difference between the various dataset versions obtained by the splitting search algorithm. Thus, we choose the first version of PETA\textsubscript{ZS} and RAP\textsubscript{ZS} as the final version for the following experiments. We illustrate the comparison between PETA and PETA\textsubscript{ZS} as an example in Fig. \ref{fig:overlap_dataset}. For the existing dataset PETA, 26.91 percent of pedestrian identities of the test set appear in the training set and occupy 55.27 percent of images in the training set. In contrast, for the proposed PETA\textsubscript{ZS}, no pedestrian identity of the test set appears in the training set.


\begin{table*}[t]
\renewcommand{\arraystretch}{1.3}
\caption{Details of existing datasets and proposed datasets. Zero-shot setting of pedestrian identities is considered in the proposed PETA\textsubscript{$zs$} and RAP2\textsubscript{$zs$} datasets. $I_{train}$, $I_{valid}$, $I_{test}$ indicates the number of identities in the training set, validation set, and test set respectively. Pedestrian identities are not provided in RAPv1, PA100k and are partly provided in RAPv2, so the exact quantity cannot be counted, which is denoted by -- . Due to the overlapped identities between $I_{train}$, $I_{valid}$, and $I_{test}$ of PETA, the sum of $I_{train}$, $I_{valid}$, and $I_{test}$ is not equal to that in PETA\textsubscript{$zs$}. \textit{Attribute} here denotes the number of attributes used for evaluation. Due to no apparent difference between the various versions, the first version of PETA\textsubscript{ZS} and RAP\textsubscript{ZS} is selected as \textbf{the final version}, which is represented by red font.}
\label{tab:five_version}
\centering
\begin{tabular}{cc|c|c|c|c|c|c|c|c|c|c|c|c|c|c}
\toprule
     \multicolumn{2}{c|}{Dataset} & Attr & $I_{all}$ & $I_{train}$ & $I_{valid}$ & $I_{test}$ & $N_{all}$ & $N_{train}$ & $N_{val}$ & $N_{test}$ & mA & Accu & Prec & Recall & F1\\ \hline \hline
    \multirow{5}{*}{PETA\textsubscript{ZS}} & \bf{v1} & \multirow{5}{*}{35} & \multirow{5}{*}{8,699} & \bf{5,233} & \bf{1,760} & \bf{1,706} &\multirow{5}{*}{19,000} & \bf{11,241} & \bf{3,826} & \bf{3,933} & \bf{69.14} & \bf{58.43} & \bf{76.65} & \bf{67.90} & \bf{71.45} \\
      & v2 &  &  & 5,212 & 1,698 & 1,789  & & 11,545 & 3,589 & 3,866 & 68.41 & 57.01 & 75.00 & 67.72 & 70.64 \\
      & v3 &  &  & 5,189 & 1,734 & 1,776  & & 11,589 & 3,584 & 3,827 & 68.25 & 58.49 & 75.74 & 68.90 & 71.62 \\
      & v4 &  &  & 5,184 & 1,713 & 1,802  & & 11,117 & 3,890 & 3,993 & 69.17 & 58.23 & 76.52 & 68.08 & 71.45 \\
      & v5 &  &  & 5,248 & 1,676 & 1,775  & & 11,720 & 3,509 & 3,771 & 68.08 & 59.33 & 77.14 & 69.20 & 72.42 \\ \cline{1-16}
     \multirow{5}{*}{RAP\textsubscript{ZS}} & \bf{v1} & \multirow{5}{*}{53} & \multirow{5}{*}{2,590} & \bf{1,566} & \bf{505} & \bf{518} & \multirow{5}{*}{26,638} &  \bf{17,062} & \bf{4,648} & \bf{4,928} & \bf{68.73} & \bf{64.84} & \bf{81.35} & \bf{74.24} & \bf{77.19}\\
      & v2 &  &  & 1,517 & 551 & 521 & &  15,175 & 5,832 & 5,631 & 67.81 & 63.79 & 80.87 & 73.35 & 76.45 \\
      & v3 &  &  & 1,601 & 514 & 474 & &  16,455 & 5,214 & 4,969 & 69.49 & 64.47 & 80.74 & 74.21 & 76.89 \\
      & v4 &  &  & 1,556 & 531 & 502 & &  15,879 & 5,447 & 5,312 & 69.99 & 64.50 & 81.28 & 73.91 & 76.96 \\
      & v5 &  &  & 1,593 & 503 & 493 & &  16,515 & 5,084 & 5,039 & 69.71 & 64.83 & 81.18 & 74.35 & 77.17 \\
    
\bottomrule
\end{tabular}
\vspace{-1em}
\end{table*}

\section{Methods} \label{method}
In this section, we start by formalizing the pedestrian attribute recognition task. Then we review and reimplement four existing state-of-the-art methods MsVAA \cite{sarafianos2018deep}, VAC \cite{guo2019visual}, ALM \cite{tang2019Improving}, and JLAC \cite{tan2020relation} on the same experimental settings to make a fair comparison. Finally, considering the enormous gap between baseline performance of various methods, we propose a solid and reliable baseline method. 

\subsection{Task Formulation and Baseline Method}
Given a training set of $\mathcal{D} = \{ (x_{i}, y_{i}), i=1,2,\dots,N \}$, where $y_{i} \in \{ 0, 1 \}^{M}$ is a binary vector with $M$ number of attributes and $y_{ij} = 1$ indicates the presence of the $j$-th attribute in $i$-th image, the task is to learn $f:X \rightarrow Y$ by minimizing the empirical risk loss:

\begin{equation}
    Loss = \frac{1}{N} \sum_{i=1}^N \mathcal{L} (y_{i}, \mathcal{C}(\mathcal{F}(x_{i};W_{f}); W_{c})),
\end{equation}
where $\mathcal{L}(\cdot)$, $\mathcal{F}(\cdot;W_{f})$, and $\mathcal{C}(\cdot;W_{c})$ denotes the loss function, feature extractor, and classifier respectively. $W_{f}$ and $W_{c}$ are the parameters in feature extractor and classifier. At the test time, the output logits is followed by a sigmoid function to get the final prediction:

\begin{align}
    \hat{y}_{ij} &= 
    \begin{cases}
      1, &  p_{ij} >= t_{cls} \\
      0, & p_{ij} < t_{cls}
    \end{cases}  \\
     p_{i} &= \sigma (\mathcal{C}(\mathcal{F}(x_{i};W_{f}); W_{c})),
\end{align}
where $\sigma(z) = 1/(1+ e^{-z})$ is the sigmoid function and classification threshold is set to $t_{cls} = 0.5$ by default. The output probability vector is denoted as $p_{i} = \{ p_{ij}|j=1, \dots, M \} \in R^{M}$, where $p_{ij} \in [0, 1]$ indicates the probability that $j$-th attribute appears in $i$-th image.

For the baseline method, we use binary cross-entropy loss as our loss function, which can be formulated as: 
\begin{align}
	\mathcal{L} & = - \sum_{j=1}^{M} \omega_{j}(y_{ij} \log(p_{ij} + (1 - y_{ij})\log(1 - p_{ij}))
	\label{eq:celoss}
\end{align}
where $\omega_{j}$ is the attribute weight of $j$-th attribute to alleviate the distribution imbalance between attributes. We find experimentally that weight functions play an important role in model optimization. Thus, we introduce three widely used weight functions for pedestrian attribute recognition. The most popular weight function proposed by Li \etal \cite{li2015deepmar} is computed as:
\begin{align}
	\omega_{j} = 
    \begin{cases}
      e^{1 - r_{j}}, &  y_{ij}=1 \\
      e^{r_{j}}, & y_{ij}=0
    \end{cases}
    \label{eq:weight1}
\end{align}
where $r_{j} $ is the positive sample ratio of $j$-$th$ attribute in the training set. Recently, Tan \etal \cite{tan2020relation} introduces a weight function, which is formulated as: 
\begin{align}
	\omega_{j} = 
    \begin{cases}
      \sqrt{\frac{1}{2r_{j}}}, &  y_{ij}=1 \\
      \sqrt{\frac{1}{2(1-r_{j})}}, & y_{ij}=0
    \end{cases}
    \label{eq:weight2}
\end{align}
Besides, to solve the general long-tailed distribution problem, Zhang \etal \cite{zhang2021disalign} proposes a weight function, which can be computed as:
\begin{align}
	\omega_{j} = 
    \begin{cases}
      \frac{(\frac{1}{r})^{\alpha}}{(\frac{1}{r})^{\alpha} + (\frac{1}{1 - r})^{\alpha}}, &  y_{ij}=1 \\
      \frac{(\frac{1}{1 - r})^{\alpha}}{(\frac{1}{r})^{\alpha} + (\frac{1}{1 - r})^{\alpha}}, & y_{ij}=0
    \end{cases}
    \label{eq:weight3}
\end{align}
where $\alpha$ is a hyper-parameter to adjust the weight between positive ratio and negative ratio. We name the weight function in Equation.~\ref{eq:weight1}, Equation.~\ref{eq:weight2}, and Equation.~\ref{eq:weight3} as WF1, WF2, and WF3 respectively. Ablation experiments of weight functions can be found in Table \ref{tab:ablation_weight}.

\subsection{Evaluated Methods} \label{eva_methods}
For making a fair comparison among state-of-the-art methods and our baseline method, we reimplement MsVAA \cite{sarafianos2018deep}, VAC \cite{guo2019visual}, ALM \cite{tang2019Improving}, and JLAC \cite{tan2020relation} methods. There are two reasons why we choose these methods to reimplement. First, all four methods do not adopt human key points \cite{li2018pose, zhao2018grouping}, human parsing \cite{li2019pedestrian, tan2019attention}, or extra labeled data \cite{tan2019attention} as auxiliary knowledge. The primary purpose of this work is to provide a reliable and solid evaluation from the perspective of existing datasets and state-of-the-art methods. Therefore, whether to use auxiliary knowledge and how to choose a pre-trained model \cite{kalayeh2018human, zhao2017spindle, zhao2017pyramid, zhang2020correlating} for auxiliary knowledge is beyond the scope of this work. Second, the authors of these methods provide detailed implementation information to achieve comparable performance compared with performance reported in the paper.


\noindent \textbf{MsVAA} \cite{sarafianos2018deep} combines attention mechanism with conv4 and conv5 stage of ResNet101 \cite{he2016deep} as the feature extractor to extract attribute-wise spatial attention map and average pooled image feature. After one image feature and two attribute features are sent as input to the linear classifier, three logits are aggregated in prediction level to get the final results. A weighted focal loss \cite{lin2017focal} is used as the loss function.


\noindent \textbf{VAC} \cite{guo2019visual} considers the human prior that the attention area of augmented samples should be consistent. For example, compared with the attention area of the original image, the attention area of horizontally flipped image should also be flipped horizontally. Thus, the feature extractor (ResNet50 \cite{he2016deep}) takes a pair of original pictures and augmented pictures as input and aligns the attention region of corresponding feature maps. The classifier takes the feature vector of the original image as input and outputs the logits as the final prediction. The weighted sigmoid cross-entropy loss in Equation \ref{eq:celoss} and Equation \ref{eq:weight1} is used as the loss function.

\noindent \textbf{ALM} \cite{tang2019Improving} draws on the idea of feature pyramid networks (FPN) \cite{lin2017feature} on object detection and integrated spatial transformation network (STN) \cite{jaderberg2015spatial} and squeeze-and-excitation \cite{hu2018squeeze} with each layer of FPN as the feature extractor. Followed by spatial transformation, each layer in FPN outputs the feature vector for each attribute. In the training stage, deep supervision is adopted in each layer of FPN. In the test stage, predictions from each layer are aggregated by element-wise maximum. The weighted sigmoid cross-entropy loss is used as in Equation \ref{eq:celoss} and Equation \ref{eq:weight1}.

\noindent \textbf{JLAC} \cite{tan2020relation} proposes a two-branch network consisted of an attribute relation graph and a contextual relation graph as the feature extractor to represent the attribute feature and image feature. Attribute relation branch learns each attribute feature and conducts feature propagation by graph convolution network \cite{kipf2017semi} to refine attribute features. The refined features are sent to one classifier. Contextual relation branch projects feature map into node vectors through NetVLAD \cite{arandjelovic2016netvlad} and concatenates the node vectors as the inputs to another classifier. The weighted sigmoid cross-entropy loss is used as in Equation \ref{eq:celoss} and Equation \ref{eq:weight2}.

\section{Experiments}

In this section, we first give an exhaustive introduction to datasets, evaluation metrics, and implementation details adopted by our experiments. Then, experimental performance on existing datasets and proposed datasets are present to make a fair comparison between our baseline method and state-of-the-art methods. Experiments comprise following three parts. First, we compare the performance of the baseline model of state-of-the-art methods to uncover the inconsistent baseline problem. Second, we sequentially conduct experiments in existing datasets and proposed datasets to compare our baseline method with state-of-the-art methods. Finally, we conduct several ablation studies on our baseline method to demonstrate the effect of various factors. 

\begin{table*}[ht]
\renewcommand{\arraystretch}{1.3}
\caption{Details of existing datasets and proposed datasets. $I_{all}$, $I_{train}$, $I_{valid}$, $I_{test}$ indicates the number of identities in the whole set, train set, validation set and test set respectively. $N_{all}$, $N_{train}$, $N_{valid}$, $N_{test}$ indicates the number of images in the whole set, train set, validation set and test set respectively. Pedestrian identities are not provided in RAP1, PA100k and are partly provided in RAP2, so the exact quantity cannot be counted, which is denoted by -- . Due to the overlapped identities between $I_{train}$, $I_{valid}$ and $I_{test}$ in PETA, $I_{all} \neq I_{train} + I_{valid} + I_{test}$. \textit{Attr} here denotes the number of attributes used for evaluation.}
\label{tab:dataset_info}
\centering
\begin{tabular} {c|c|c|c|c|c|c|c|c|c|c}
\toprule
    Setting & Dataset & Attr & $I_{all}$ & $I_{train}$ & $I_{valid}$ & $I_{test}$ & $N_{all}$ & $N_{train}$ & $N_{val}$ & $N_{test}$ \\ \hline \hline
    \multirow{3}{*}{random splitting} & PETA & 35 & 8,699 & 4,886 & 1,264  & 4,110  &  19,000 & 9,500 & 1,900 & 7,600 \\ 
    & RAP1 & 51 & -- & -- & -- & -- & 41,585 & 33,268 & -- & 8,317 \\ 
    & RAP2 & 54 & -- & -- & -- & --  & 84,928 & 50,957 & 16,986 & 16,985 \\ \cline{1-1}
    \multirow{3}{*}{zero-shot splitting}& PETA\textsubscript{ZS} & 35 & 8,699 & 5,233 & 1,760 & 1,706 & 19,000 & 11,241 & 3,826 & 3,933 \\
    & RAP\textsubscript{ZS} & 53 & 2,590 & 1,566 & 505 & 518 & 26,632 &  17,062 & 4,648 & 4,928 \\
    & PA100k & 26 & -- & -- & -- & -- & 100,000 & 80,000 & 10,000 & 10,000  \\ 
\bottomrule
\end{tabular}
\vspace{-1em}
\end{table*}

\subsection{Datasets and Evaluation Metrics}

We conduct experiments in six datasets, including four existing datasets PETA \cite{deng2014pedestrian}, RAP1 \cite{li2016richly}, RAP2 \cite{li2018richly}, PA100K \cite{liu2017hydraplus}, and two proposed datasets PETA\textsubscript{ZS}, RAP\textsubscript{ZS}. We divide datasets into two categories according to the splitting criteria. Some are random-splitting datasets, where samples are random assigned to the training set and test set, \ie, PETA, RAP1, RAP2. Others are zero-shot splitting datasets, where samples are assigned according to the zero-shot pedestrian identities of the training set and test set, \ie, PETA\textsubscript{ZS}, RAP\textsubscript{ZS}, and PA100K. The details are given in Table \ref{tab:dataset_info}. For PETA, Deng \etal annotates each image with more than 60 attributes, but only 35 attributes are used for evaluation due to the imbalance distribution of some attributes. For RAP and RAP2, Li \etal annotates each image with 72 attributes, but only 51 and 54 attributes are respectively selected for evaluation considering the proportion of positive samples. For PA100K, Liu \etal annotates each image with 26 attributes and adopts all attributes for training and testing. We find experimental that training with all annotated attributes cannot consistently improve the performance. Thus, we only choose 35, 51, 54 attributes for training on PETA, RAP1, RAP2 respectively.

According to the previous works \cite{liu2017hydraplus, sarafianos2018deep, tang2019Improving, guo2019visual, li2018richly, tan2019attention}, we adopt five metrics to evaluate the model performance, including four instance-level metrics and one attribute-level (label-level) metric. For instance-level metrics, accuracy (Acc), precision (Prec), recall (Recall), F1 are adopted and calculated as follows:
\begin{align}
    Acc &= \frac{1}{N} \sum_{i=1}^N \frac{TP_{i}}{TP_{i} + FP_{i} + FN_{i}} \\
    Prec &= \frac{1}{N} \sum_{i=1}^N \frac{TP_{i}}{TP_{i} + FP_{i}} \\
    Recall &= \frac{1}{N} \sum_{i=1}^N \frac{TP_{i}}{TP_{i} + FN_{i}} \\
    F1 &= \frac{1}{N} \sum_{i=1}^N \frac{2 \cdot Prec \cdot Recall}{Prec + Recall} \label{eq:f1}
\end{align}
where $TP_{i}$, $FP_{i}$, $FN_{i}$ is the number of true positive, false positive, false negative attributes of $i$-th sample and $N$ denotes the number of samples. For attribute-level metrics, mean accuracy (mA) is adopted and calculated as follows:
\begin{align}
    mA &= \frac{1}{M} \sum_{j=1}^M \frac{1}{2} (\frac{TP^{j}}{TP^{j} + FN^{j}} + \frac{TN^{j}}{TN^{j} + FP^{j}}) \label{eq:ma_1}
\end{align}
where $TP^{j}$, $TN^{j}$, $FP^{j}$, $FN^{j}$ is the number of true positive, true negative, false positive, false negative samples of $j$-th attribute and $M$ denotes the number of attributes.

\subsection{Implementation Details} \label{implement}

To make a fair comparison, all experiments of reimplemented methods and the proposed baseline method are conducted on the same experimental settings. For previous state-of-the-art methods, we strictly follow their papers and public codes to reimplement corresponding models. All models are implemented with PyTorch and trained end-to-end. ResNet50 is adopted as the backbone network. Pedestrian images are resized to 256 × 128 (height × width) as inputs. We use Adam optimizer to train the whole network with initial learning rate 1e-3 and weight decay 5e-4. The ReduceLROnPlateau learning rate scheduler is adopted to reduce the learning rate with 0.1 factor once the performance has no improvement in four epochs. For the JLAC method, since the Batch Normalization \cite{ioffe2015batch} is used after the classification layer, we use the MultiStepLR scheduler with milestones [10, 20] to adjust the learning rate. The batch size is set to 64, and the total epoch number of training stage is set to 30. Only image resizing, random horizontal flipping, and normalization are used as augmentation at the training time. Only image resizing is adopted during test. The weighting function \ref{eq:weight2} is used by default.

We emphasize that our experiments do not focus on achieving higher performance on the metrics but show how good a simple baseline model can achieve and how big the gap between the state-of-the-art method and the baseline model is.

\subsection{Comparison of our baseline method with those in state-of-the-art methods}

\begin{table*}[ht]
\renewcommand{\arraystretch}{1.3}
\caption{Performance comparison of \textbf{baseline models} of state-of-the-art methods on the PETA, RAP1, PA100k datasets. Performance on five metrics, mA, Accuracy, Precision, Recall, and F1 are evaluated. Although some baseline models use the same backbone network, their performance varies greatly.}
\label{tab:baseline}
\centering
\resizebox{\linewidth}{!}{
\begin{tabular}{c|c|ccccc|ccccc|ccccc}
\toprule
\multirow{2}{*}{Method} & \multirow{2}{*}{Backbone} & \multicolumn{5}{c|}{PETA} & \multicolumn{5}{c|}{RAP} &\multicolumn{5}{c}{PA100K} \\ \cline{3-17} 
	&	& mA & Accu & Prec & Recall & F1 & mA & Accu & Prec & Recall & F1 & mA & Accu & Prec & Recall & F1 \\ \hline \hline
GRL\cite{zhao2018grouping} (IJCAI18) & Inception-V3 & 81.50 & -- & 89.70 & 81.90 & 85.68 & 76.10 & -- & 82.20 &  74.80 & 78.30 & -- & -- & -- & -- & -- \\
VRKD\cite{li2019pedestrian} (IJCAI19) & ResNet50 & 81.27 & 76.69 & 87.33 & 82.76 & 84.99 & 75.12 & 66.67 & 81.16 & 76.52 & 79.00 & 76.31 & 76.76 & 88.62 & 83.22 & 85.84 \\
AAP\cite{han2019attribute} (IJCAI19) & ResNet50 & 84.68 & 78.89 & 85.38 & 86.41 & 85.83 & 79.67 & 66.01 & 77.93 & 79.25 & 78.58 & 78.12 & 74.11 & 84.42 & 84.09 & 84.25 \\ 
MsVAA\cite{sarafianos2018deep} (ECCV18) & ResNet101 & 82.67 & 76.63 & 85.13 & 84.46 & 84.79 & -- & -- & -- & -- & -- & -- & -- & -- & -- & -- \\
VAC \cite{guo2019visual} (CVPR19) & ResNet50 & -- & -- & -- & -- & -- & -- & -- & -- & -- & -- & 78.12 & 75.23 & 88.47 & 83.41 & 85.86\\
ALM\cite{tang2019Improving} (ICCV19) & BN-Inception & 82.66 & 77.73 & 86.68 & 84.20 & 85.57 & 75.76 & 65.57 & 78.92 & 77.49 & 78.20 & 77.47 & 75.05 & 86.61 & 85.34 & 85.97 \\ 
JLAC\cite{tan2020relation} (AAAI20) & BN-Inception & 84.81 & 77.66 & 85.43 & 85.85 & 85.64 & 80.49 & 66.00 & 76.35 & 81.32 & 78.76 & 81.17 & 77.84 & 85.95 & 87.19 & 86.56 \\ \hline
Baseline & BN-Inception & 83.34 & 76.11 & 85.21 & 83.89 & 84.55 & 78.66 & 65.55 & 77.87 & 78.71 & 78.28 & 79.73 & 77.68 & 86.61 & 86.29 & 86.45 \\
Baseline & ResNet50 & 84.42 & 78.13 & 86.88 & 85.08 & 85.97 & 80.32 & 67.28 & 79.04 & 79.89 & 79.46 & 80.38 & 78.58 & 87.09 & 87.01 & 87.05 \\
Baseline & ResNet101 & 85.17 & 78.82 & 86.77 & 86.11 & 86.44 & 80.82 & 67.56 & 79.01 & 80.37 & 79.68 & 81.61 & 79.45 & 87.66 & 87.59 & 87.62 \\\bottomrule
\end{tabular}}
\vspace{-1em}
\end{table*}

A solid baseline method is essential for developing specific area, such as image classification \cite{he2019bag}, person re-identification \cite{luo2019strong}, and zero-shot learning \cite{xian2018zero}.  However, baseline model performance of state-of-the-art methods in pedestrian attribute recognition varies greatly, which hinders fair comparison between methods and cannot fully validate the method contribution. As listed in Table \ref{tab:baseline}, there is a significant performance gap between existing methods in each metric, especially in mA. For PETA, RAP, PA100K, performance in mA fluctuates by as much as 3.41, 4.55, 1.81 points, which is a considerable performance variance compared with the performance improvement from the technical contribution. Except for JLAC \cite{tan2020relation}, which adopts horizontal flipping, random scaling, rotation, translation, cropping, erasing, and adding random gaussian blurs as data augmentations, our baseline method achieves performance superior to other baseline models, with only horizontal flipping augmentation. For example, compared with ALM \cite{tang2019Improving}, our baseline model with BN-Inception achieve 0.68, 2.90, 2.26 performance improvement in mA and 0.40, 1.26, 1.08 performance improvement in F1. We argue that the proposed solid baseline without bells and whistles plays an important role in advancing the pedestrian attribute recognition field by providing a performance reference to evaluate the model technical contribution.

\begin{table*}[ht]
\renewcommand{\arraystretch}{1.3}
\caption{Performance comparison of state-of-the-art methods on the PETA, RAPv1, PA100k datasets. Performance on five metrics, mA, Accuracy, Precision, Recall, and F1 are evaluated. Parameters (Params) and floating point operations per second operations (FLOPS) of various methods are also reported.}
\label{tab:sota_perf}
\centering
\resizebox{\linewidth}{!}{
\begin{tabular}{c|c|ccccc|ccccc|ccccc|cc}
\toprule
\multirow{2}{*}{Method} & \multirow{2}{*}{Backbone} & \multicolumn{5}{c|}{PETA} & \multicolumn{5}{c|}{RAP1} &\multicolumn{7}{c}{PA100k} \\ \cline{3-19} 
	&	& mA & Accu & Prec & Recall & F1 & mA & Accu & Prec & Recall & F1 & mA & Accu & Prec & Recall & F1 & Params(M) & FLOPS(G) \\ \hline \hline
DeepMAR \cite{li2015deepmar} (ACPR15) & CaffeNet & 82.89 & 75.07 & 83.68 & 83.14 & 83.41 & 73.79 & 62.02 & 74.92 & 76.21 & 75.56 & 72.70 & 70.39 & 82.24 & 80.42 & 81.32  & -- & --\\
HPNet\cite{liu2017hydraplus} (ICCV17) & InceptionNet & 81.77 & 76.13 & 84.92 & 83.24 & 84.07 & 76.12 & 65.39 & 77.33 & 78.79 & 78.05 & 74.21 & 72.19 & 82.97 & 82.09 & 82.53  &--&--\\
JRL \cite{wang2017attribute} (ICCV17) \footnotemark[1] & AlexNet & 85.67 & -- & 86.03 & 85.34 & 85.42 & 77.81 & -- & 78.11 & 78.98 & 78.58 &--&--&--&--&--& -- & -- \\
JRL \cite{wang2017attribute} (ICCV17) & AlexNet & 82.13 & -- & 82.55 & 82.12 & 82.02 & 74.74 & -- & 75.08 & 74.96 & 74.62 &--&--&--&--&-- & -- & -- \\
LGNet \cite{liu2018localization} (BMVC18) & Inception-V2 &--&--&--&--&--& 78.68 & 68.00 & 80.36 & 79.82 & 80.09 & 76.96 & 75.55 & 86.99 & 83.17 & 85.04  &--&--\\
PGDM \cite{li2018pose} (ICME18) & CaffeNet & 82.97 & 78.08 & 86.86 & 84.68 & 85.76 & 74.31 & 64.57 & 78.86 & 75.90 & 77.35 & 74.95 & 73.08 & 84.36 & 82.24 & 83.29 &--&--\\ \hline
GRL\cite{zhao2018grouping} (IJCAI18) & Inception-V3 & 86.70 & -- & 84.34 & \bf{88.82} & 86.51 & 81.20 & -- & 77.70 & 80.90 & 79.29 &--&--&--&--&-- & -- &--\\
RA\cite{zhao2019recurrent} (AAAI19)& Inception-V3 & 86.11 & -- & 84.69 & 88.51 & 86.56 & 81.16 & -- & 79.45 & 79.23 & 79.34 &--&--&--&--&--& -- & -- \\
VSGR\cite{li2019visual} (AAAI19)\footnotemark[1] & ResNet50 & 85.21 & \bf{81.82} & \bf{88.43} & 88.42 & \bf{88.42} & 77.91 & \bf{70.04} & 82.05 & 80.64 & \bf{81.34} & 79.52 & \bf{80.58} & 89.40 & 87.15 & \bf{88.26} & -- & --\\
VRKD\cite{li2019pedestrian} (IJCAI19)& ResNet50 & 84.90 & 80.95 & 88.37 & 87.47 & 87.91 & 78.30 & 69.79 & 82.13 & 80.35 & 81.23 & 77.87 & 78.49 & 88.42 & 86.08 & 87.24 & -- & -- \\
AAP\cite{han2019attribute} (IJCAI19) & ResNet50 & \bf{86.97} & 79.95 & 87.58 & 87.73 & 87.65 & 81.42 & 68.37 & 81.04 & 80.27 & 80.65 & 80.56 & 78.30 & \bf{89.49} & 84.36 & 86.85  & -- & -- \\ \hline
MsVAA\cite{sarafianos2018deep} (ECCV18) & ResNet101 & 84.59 & 78.56 & 86.79 & 86.12 & 86.46 & -- & -- & -- & -- & -- & -- & -- & -- & -- & -- & -- & --\\
VAC \cite{guo2019visual} (CVPR19) & ResNet50 & -- & -- & -- & -- & -- & -- & -- & -- & -- & -- & 79.16 & 79.44 & 88.97 & 86.26 & 87.59 & -- & -- \\
ALM\cite{tang2019Improving} (ICCV19) & BN-Inception & 86.30 & 79.52 & 85.65 & 88.09 & 86.85 & \bf{81.87} & 68.17 & 74.71 & \bf{86.48} & 80.16 & 80.68 & 77.08 & 84.21 & 88.84 & 86.46 & -- & -- \\ 
JLAC\cite{tan2020relation} (AAAI20) & BN-Inception & 86.96 & 80.38 & 87.81 & 87.09 & 87.45 & 83.69 & 69.15 & 79.31 & 82.40 & 80.82 & \bf{82.31} & 79.47 & 87.45 & \bf{87.77} & 87.61 & -- & -- \\ \hline
MsVAA\cite{sarafianos2018deep} (ECCV18)  \footnotemark[2] & ResNet50 & 83.57 & 77.62 & 86.74 & 84.57 & 85.64 & 78.86 & 66.99 & 79.39 & 79.15 & 79.27 & 80.41 & 78.37 & 87.08 & 86.52 & 86.80 & 114.19 & 3.28\\
VAC \cite{guo2019visual} (CVPR19)  \footnotemark[2] & ResNet50 & 84.58 & 78.37 & 87.40 & 84.93 & 85.83 & 80.27 & 66.26 & 77.87 & 79.77 & 78.36 & 80.11 & 78.43 & 87.24 & 86.44 & 86.45 & 23.56 & 2.69 \\  
ALM\cite{tang2019Improving} (ICCV19) \footnotemark[2] & ResNet50 & 85.57 & 76.91 & 83.48 & 87.08 & 85.24 & 80.96 & 66.04 & 75.13 & 82.69 & 78.73 & 81.29 & 77.61 & 85.42 & 87.40 & 86.40 & 27.70 & 2.82 \\ 
JLAC\cite{tan2020relation} (AAAI20) \footnotemark[2] & ResNet50 & 86.88 & 78.03 & 85.61 & 86.21 & 85.91 & 81.51 & 66.29 & 77.17 & 80.42 & 78.76 & 82.10 & 77.49 & 85.74 & 86.88 & 86.31 & 40.87 & 6.00 \\ 
\hline
Baseline & BN-Inception & 83.34 & 76.11 & 85.21 & 83.89 & 84.55 & 78.66 & 65.55 & 77.87 & 78.71 & 78.28 & 79.73 & 77.68 & 86.61 & 86.29 & 86.45 & \bf{11.32} & \bf{1.34}\\
Baseline & ResNet50 & 84.42 & 78.13 & 86.88 & 85.08 & 85.97 & 80.32 & 67.28 & 79.04 & 79.89 & 79.46 & 80.38 & 78.58 & 87.09 & 87.01 & 87.05 & 23.56 & 2.69\\
Baseline & ResNet101 & 85.17 & 78.82 & 86.77 & 86.11 & 86.44 & 80.82 & 67.56 & 79.01 & 80.37 & 79.68 & 81.61 & 79.45 & 87.66 & 87.59 & 87.62 & 42.55 & 5.12\\\bottomrule
\end{tabular}}
\vspace{-1em}
\end{table*}

\subsection{Comparison of our baseline method with state-of-the-art methods on existing datasets}

We make a systematic performance comparison between our baseline methods with state-of-the-art algorithms on the PETA, RAP1, and PA100K in Table \ref{tab:sota_perf}. For a fair comparison, besides the performance reported by the papers \cite{sarafianos2018deep, guo2019visual, tang2019Improving, tan2020relation}, we also report the performance of our reimplementations based on the same experimental settings described in Implementation Details. Model complexity is also reported in parameters (Params) and floating point operations per second operations (FLOPS). In order to give a clear comparison and avoid the effect of overfitting, all experiments are conducted only with resizing, random horizontal flipping, and normalization. Considering that the F1 is a trade-off between Prec and Recall, we argue that F1 and mA are the two most important metrics for method evaluation.

Our reimplemented methods achieve comparable performance compared with the performance reported by the paper. For MsVAA \cite{sarafianos2018deep} , our reimplemented method achieves slightly worse performance because we adopt ResNet50 as the backbone network instead of ResNet101 \footnote{\url{https://github.com/cvcode18/imbalanced_learning}} in the original paper. For VAC \cite{guo2019visual} \footnote{\url{https://github.com/hguosc/visual_attention_consistency}}, our reimplemented method achieves comparable performance on PA100K, namely 0.95 performance improvement in mA and 1.14 performance decrease in F1. For ALM \cite{tang2019Improving} \footnote{\url{https://github.com/chufengt/iccv19_attribute}}, our reimplemented method achieves slightly lower performance than paper-reported performance on PETA and RAP1 and achieves comparable even higher performance on PA100K. There are two reasons. First, the authors of ALM adopt a two-stage training manner and use a warm-up learning rate scheduler, but we train the model once end-to-end to align with other methods. Second, due to the extremely similar images between the training set and test set on PETA and RAP, caused by random splitting and shown in Figure \ref{fig:zeroshot_overview}, models of PETA and RAP incline to overfitting. This phenomenon can also be found in JLAC experiments. For JLAC, since authors adopt a series of data augmentations described in implementation details, our reimplemented method can achieve comparable performance on PA100K and slightly worse performance on PETA and RAP1. Thus, we conclude that, due to the small size and highly similar images between the training set and test set, which make models easy to overfit, performance on PETA and RAP1 is more sensitive to data augmentations than performance on PA100K. We verify this conclusion in Table \ref{tab:ablation_aug}.

After achieving the comparable performance of reimplemented methods, we can directly compare our baseline methods with state-of-the-art methods in the same experimental settings. We use the baseline method with ResNet50 by default. As listed in Table \ref{tab:sota_perf}, compared with reimplementations of MsVAA and VAC, our baseline method achieve comparable even higher performance on three datasets. Compared with ALM, on three datasets, our baseline method with only 85\% parameters achieves performance improvement in F1 by 0.65 to 0.73, while mA performance reduces by 0.64 to 1.15. Compared with JLAC, the baseline performance is reduced by 1.56, 1.19, 1.72 in mA on PETA, RAP1, and PA100K separately and improved by 0.80, 0.70, 0.74 in F1. Although performance decrease on mA is higher than the performance increase on F1, our baseline only uses 57.65\% parameters and 44.83\% flops compared with JLAC, which uses the two-branch network and each branch adopts conv5 stage of ResNet50 as unshared parameters.

\footnotetext[1]{Results are achieved by the ensemble model.}
\footnotetext[2]{Results are reimplemented with the same experimental setting of our baseline.}

\begin{table*}[ht]
\renewcommand{\arraystretch}{1.3}
\caption{Performance comparison of four methods on the PETA, RAP2 datasets. We use \textit{``zero-shot"} to denote our proposed PETA\textsubscript{$zs$} or RAP\textsubscript{$zs$} dataset, while \textit{``random"} indicates the existing datasets PETA or RAP. Performance on five metrics, mA, Accuracy, Precision, Recall, and F1 are evaluated. Our baseline method adopt ResNet50 as the backbone network.}
\label{tab:proposed_dataset_perf}
\centering
\begin{tabular}{c|c|ccccc|ccccc}
\toprule
\multirow{2}{*}{Method} & \multirow{2}{*}{Dataset Splitting} & \multicolumn{5}{c|}{PETA} & \multicolumn{5}{c}{RAP2} \\ \cline{3-12} 
	&	& mA & Accu & Prec & Recall & F1 & mA & Accu & Prec & Recall & F1  \\ \hline \hline
\multirow{2}{*}{MsVAA} & random & 84.35 & 78.69 & 87.27 & 85.51 & 86.09 & 78.34 & 65.57 & 77.37 & 79.17 & 78.26 \\ \cline{2-2}
 & zero-shot & 71.53 & 58.67 & 74.65 & 69.42 & 71.94 & 72.04 & 62.13 & 75.67 & 75.81 & 75.74\\\cline{1-12} 
\multirow{2}{*}{VAC} & random & 83.63 & 78.94 & 87.63 & 85.45 & 86.23 & 79.23 & 64.51 & 75.77 & 79.43 & 77.10 \\\cline{2-2}
 & zero-shot & 71.91 & 57.72 & 72.05 & 70.64 & 70.90 & 73.70 & 63.25 & 76.23 & 76.97 & 76.12\\\cline{1-12} 
\multirow{2}{*}{ALM} & random & 85.11 & 79.14 & 86.99 & 86.33 & 86.39 & 79.79 & 64.79 & 73.93 & 82.03 & 77.77 \\\cline{2-2} 
 & zero-shot & 73.01 & 57.78 & 69.50 & 73.69 & 71.53 & 74.28 & 63.22 & 72.96 & 80.73 & 76.65\\\cline{1-12} 
\multirow{2}{*}{JLAC} & random & 86.02 & 79.51 & 86.62 & 87.19 & 86.66 & 79.23 & 64.42 & 75.69 & 79.18 & 77.40 \\\cline{2-2}
 & zero-shot & 73.60 & 58.66 & 71.70 & 72.41 & 72.05 & 76.38 & 62.58 & 73.14 & 79.20 & 76.05\\ \cline{1-12} 
 \multirow{2}{*}{Baseline} & random & 84.42 & 78.13 & 86.88 & 85.08 & 85.97 & 79.19 & 65.27 & 76.26 & 79.89 & 78.03 \\\cline{2-2}
 & zero-shot & 71.62 & 58.19 & 73.09 & 70.33 & 71.68 & 72.32 & 63.61 & 76.88 & 76.62 & 76.75\\\bottomrule
\end{tabular}
\vspace{-1em}
\end{table*}

\subsection{Performance comparison between existing datasets and proposed datasets}

To give a direct comparison between existing datasets (PETA, RAP2) and proposed datasets (PETA\textsubscript{ZS}, RAP\textsubscript{ZS}), we evaluate the state-of-the-art methods and our proposed baseline method on four datasets, as shown in Table \ref{tab:proposed_dataset_perf}. We emphasize that the ``random" splitting manner and ``zero-shot" splitting manner in Table \ref{tab:proposed_dataset_perf} is adopted by existing datasets (PETA, RAP2) and proposed datasets (PETA\textsubscript{ZS}, RAP\textsubscript{ZS}), separately. All experiments are performed following implementation details. We have the following observations.

First, for the same method, the performance on PETA\textsubscript{ZS} and RAP\textsubscript{ZS} decreases significantly compared with the corresponding performance on PETA and RAP2. The main reason is the pedestrian identity overlap between the training set and test set on PETA and RAP (shown in Fig. \ref{fig:common_image} and Fig. \ref{fig:overlap_dataset}), which results in high performance on common-identity images in the test set (shown in Fig. \ref{fig:overlapped_perf_gap}). However, the high performance on common-identity images does not reflect the method generalization ability. In contrast, no pedestrian identity of the test set appears in the training set on our proposed datasets, PETA\textsubscript{ZS} and RAP\textsubscript{ZS}, where the generalization ability of methods can be truly measured. We can notice that, although PETA\textsubscript{ZS} has more training samples and fewer test samples than PETA, the performance of all metrics is at least ten points lower on PETA\textsubscript{ZS} than on PETA, which demonstrates that our proposed datasets can evaluate the model generalization more rigorously and accurately. In addition, the performance on PETA\textsubscript{ZS} is close to the performance on unique-identity images of PETA test set (shown in Fig. \ref{fig:overlapped_perf_gap}), which also verifies that our proposed datsets can reflect the generalization ability of the model on unseen pedestrian images.

Second, for the same method, compared with the performance gap between PETA and PETA\textsubscript{ZS}, the performance gap between RAP2 and RAP\textsubscript{ZS} is much smaller. We consider that is related to the composition of datasets, \ie, PETA is composed of 10 publicly small-scale datasets, while RAP2 is collected in one scene. When there is no pedestrian identity overlap between the training set and test set, the cross-domain problem is inevitable. Thus, the domain difference between the training set and the test set of PETA\textsubscript{ZS} is more significant than that of RAP\textsubscript{ZS}, which results in the notable performance degradation from PETA to PETA\textsubscript{ZS}. 

Third, whether on PETA or RAP2, the performance difference between methods on the proposed dataset is more significant than the performance difference between the methods on the existing dataset. For example, five methods performance on RAP2 is from 78.34 to 79.79 in mA, while performance on RAP\textsubscript{ZS} is from 72.04 to 76.38 in mA. Because the existing datasets are prone to the overfitting problem, there is little difference in the performance of various methods, which cannot adequately represent the generalization ability of the model and hinders the direct comparison between models.

Therefore, we conclude that our proposed datasets PETA\textsubscript{ZS} and RAP\textsubscript{ZS} are superior to existing datasets PETA, RAP1, and RAP2, in terms of evaluation of the model generalization and practical application.

\subsection{Ablation Study}
In this section, we conduct experiments on several factors that affect baseline performance. Considering that PA100K is currently the largest dataset and conforms to the zero-shot setting, all experiments are conducted on PA100K and adopt ResNet50 as the backbone network. No weighting function is used. Other details are following implementation details. 

First, we make a performance comparison on the input image size. We list several common input image sizes in Table \ref{tab:input_size}. Unlike general multi-label classification on COCO \cite{lin2014microsoft} and PASCAL-VOC \cite{everingham2010pascal}, where increasing the input image size can significantly improve performance, input image size has little effect on performance. The reason is that the high-resolution ($640 \times 480$) images of COCO and PASCAL are usually resized to a small size ($224 \times 224$) as inputs. Thus, increasing input size (from 224x224 to 448x448) can bring more information and improve performance. However, almost all pedestrian images captured by surveillance cameras at far distance are low-resolution. Therefore, increasing the input image size to exceed the original image size has little effect. 

\begin{table}[t]
\caption{Performance comparison on the input image size (height x width) of baseline method on PA100K.}
\label{tab:input_size}
\centering
\begin{tabular}{c|c|c|c|c|c}
\toprule
	Input Size & mA & Accu & Prec & Recall & F1 \\\hline \hline
	224 x 224 & 78.12 & 78.60 & 89.01 & 85.13 & 86.60 \\
	256 x 128 & \bf{78.62} & \bf{79.15} & \bf{89.46} & \bf{85.49} & \bf{87.01} \\
	256 x 192 & 78.57 & 78.98 & 89.38 & 85.33 & 86.87 \\
	384 x 192 & 78.35 & 78.94 & 89.59 & 85.12 & 86.86 \\
\bottomrule
\end{tabular}
\vspace{-1em}
\end{table}

Second, we make a performance comparison on the weighting function \ref{eq:weight1}, \ref{eq:weight2}, and \ref{eq:weight3} in Table \ref{tab:ablation_weight}. Interestingly, there is almost no difference between the various weighting functions. These functions are effective in the mA metric but have little effect in the F1 metric. To explain the reason of this phenomenon, we first introduce the formulation of mA in Eq. \ref{eq:ma_1} from another perspective as follows: 

\begin{align}
    mA &= (mPR + mNR) / 2. \label{eq:ma} \\ 
    mPR &= \frac{1}{M} \sum_{1}^{M} \frac{TP^{j}}{TP^{j} + FN^{j}} = \frac{1}{M} \sum_{1}^{M} \frac{TP^{j}}{N_{j}^{p}} \label{eq:mpr}\\ 
    mNR &= \frac{1}{M} \sum_{1}^{M} \frac{TN^{j}}{TN^{j} + FP^{j}} = \frac{1}{M} \sum_{1}^{M} \frac{TN^{j}}{N_{j}^{n}} \label{eq:mnr}
\end{align}
where $N_{j}^{p}$, $N_{j}^{n}$ is the number of positive samples and negative samples of the $j$-th attribute. $mPR$ and $mNR$ indicates the mean positive recall and mean negative recall respectively.

We emphasize that all weighting functions are proposed to solve the imbalanced distribution of attributes. For most attributes, the number of negative samples is much larger than the number of positive samples $N_{j}^{n} \gg N_{j}^{p}$, which results in $TN^{j} \gg TP^{j}$. Thus, to improve the performance of positive samples of attributes,  all weighting functions pay more attention to the loss of positive samples and reduce the proportion of negative samples, which can be concluded from Eq. \ref{eq:weight1}, Eq. \ref{eq:weight2}, and Eq. \ref{eq:weight3} with $r_{j} < 0.5$. As a result, methods that adopted weighting functions can significantly increase TP at the price of decreasing TN.

For the F1 metric in Eq.~\ref{eq:f1}, which is an instance-level metric and averaged on the number of all samples, compared with the baseline performance without weighting function, methods with weighting function improve the performance Recall but achieve lower performance in Prec. Thus, considering that F1 performance is a trade-off between Prec and Recall, there is limited performance improvement in F1. On the other hand, for the mA metric, which is an attribute-level metric and averaged on the number of attributes, due to the huge difference in the denominator between Eq. \ref{eq:mpr} and Eq. \ref{eq:mnr}, the increased number of TP samples significantly improve the performance in mPR, while the decreased TN only slightly affect the performance in mNR. For example, compared with the baseline method, WF2 improves 4.18 performance at the cost of only 0.27 performance reduction, which results in 1.76 performance improvements in mA. Thus, we conclude that the model with slightly low performance in mNR can achieve much high performance in mPR, which results in performance improvement in mA.

\begin{table}[t]
\caption{Performance comparison on the weight function of baseline method on PA100K. $R_{pos}$ and $R_{neg}$ denotes the mean positive recall and mean negative recall of attribute, where mA = (mPR + mNR) / 2. We use ``--" to denote the method without weighting function. }
\label{tab:ablation_weight}
\centering
\resizebox{\linewidth}{!}{
\begin{tabular}{c|c|c|c|c|c|c|c}
\toprule
	Weight Function & mA & mPR & mNR & Accu & Prec & Recall & F1 \\\hline \hline
	-- & 78.62 & 62.98 & 94.26 & 79.15 & 89.46 & 85.49 & 87.01 \\
    WF1 & 80.19 & 66.37 & 94.01 & 78.78 & 87.86 & 86.47 & 87.16 \\
    WF2 & 80.38 & 67.16 & 93.59 & 78.58 & 87.09 & 87.01 & 87.05 \\
    WF3 $\alpha = 0.1$ & 78.93 & 63.69 & 94.17 & 79.03 & 89.04 & 85.67 & 87.32 \\
    WF3 $\alpha = 0.4$ & 80.77 & 67.75 & 93.79 & 78.82 & 87.52 & 86.83 & 87.17 \\
    WF3 $\alpha = 0.7$ & 82.44 & 72.18 & 92.70 & 75.30 & 82.70 & 87.31 & 84.94 \\
    WF3 $\alpha = 1$ & 85.05 & 78.07 & 92.02 & 75.07 & 81.40 & 88.54 & 84.82 \\
\bottomrule
\end{tabular}}
\vspace{-1em}
\end{table}

Third, we make a performance comparison on the data augmentations in Table \ref{tab:ablation_aug}, \ie, padding and random cropping, color jitter and random rotation, random erasing. Three data augmentations are sequentially added to the baseline method with ResNet50 as the backbone network. We only reported the two most important metrics mA and F1 for convenience. Compared with performance variance on PA100K, performance on PETA and RAP is more sensitive to data augmentations.

\begin{table}[ht]
\renewcommand{\arraystretch}{1.3}
\caption{Performance comparison on data augmentations. ``RC", ``CJ", ``RR" and ``RE" denotes RandomCrop, ColorJitter, RandomRotation and RandomErasing respectively.}
\label{tab:ablation_aug}
\centering
\resizebox{0.9\linewidth}{!}{
\begin{tabular}{c|cc|cc|cc|cc|cc}
\toprule
\multirow{2}{*}{Method} & \multicolumn{2}{c|}{PETA\textsubscript{ZS}} & \multicolumn{2}{c|}{PETA} & \multicolumn{2}{c|}{RAP} & \multicolumn{2}{c|}{RAP\textsubscript{ZS}} & \multicolumn{2}{c}{PA100K}\\ \cline{2-11}
& mA & F1 & mA & F1 & mA & F1 & mA & F1 & mA & F1 \\ \hline \hline
basline & 71.62 & 71.68 & 84.42 & 85.97 & 80.32 & 79.46 & 72.32 & 76.75 & 80.38 & 87.05\\\cline{1-1} 
+ Pad \& RC & 72.21 & 71.33 & 85.12 & 86.03 & 81.34 & 79.63 & 73.36 & 76.68 & 80.99 & 86.77 \\\cline{1-1} 
+ CJ \& RR & 72.55 & 70.96 & 85.00 & 85.95 & 81.33 & 79.80 & 74.14 & 77.07 & 81.38 & 87.19\\\cline{1-1} 
+ RE & 73.30 & 71.19 & 85.61 & 86.23 & 82.14 & 80.13 & 75.08 & 77.23 & 80.94 & 87.18\\ \bottomrule
\end{tabular}}
\vspace{-1em}
\end{table}

\section{Conclusion}
In this paper, we review and rethink the development of pedestrian attribute recognition from the definition, datasets, and state-of-the-art methods. For the definition of pedestrian attribute recognition, our work proposes an explicit, complete, and specific definition for the first time, which points out the essential requirement of pedestrian attribute recognition, \ie, the model can correctly predict attributes on unseen pedestrians. Guided by definition, we expose the flaws and limitations in existing datasets PETA, RAP1, and RAP2, which mislead the algorithm evaluation. The algorithm performance is overestimated and can not truly reflect the generalization ability. Thus, to solve the mentioned problems, we propose two datasets PETA\textsubscript{ZS} and RAP\textsubscript{ZS} following the zero-shot settings on pedestrian identity. Finally, we propose a solid baseline method and reimplement state-of-the-art methods to provide a reliable and realistic evaluation on four existing datasets and two proposed datasets. On the one hand, we found that our proposed baseline method, which is superior to several methods, can be taken as a reference for future work. On the other hand, due to reducing the overfitting influence, it is easier to compare methods on our proposed datasets, which directly reflect the pros and cons of various methods.

\bibliographystyle{IEEEtran}
\bibliography{main.bib}

\ifCLASSOPTIONcaptionsoff
  \newpage
\fi

\end{document}